\documentclass{article}

\usepackage{microtype}
\usepackage{graphicx}
\usepackage{subfigure}
\usepackage{booktabs} 
\usepackage{times}
\usepackage{epsfig}
\usepackage{amsmath}
\usepackage{amssymb}
\usepackage{bm}
\usepackage{amsthm}
\usepackage{multirow}
\usepackage{caption}
\usepackage[super]{nth}

\usepackage{amssymb}%
\usepackage{mathrsfs}%
\usepackage{nicefrac}
\usepackage{color,xcolor}
\usepackage{pifont}
\usepackage{array}
\usepackage{wrapfig}
\usepackage{mathtools}
\usepackage{colortbl}
\usepackage{mdframed}

\usepackage{hyperref}

\usepackage[accepted]{icml2022}

\DeclareMathOperator*{\argmax}{arg\,max}

\DeclareMathOperator{\sgn}{sign}

\newtheorem{lemma}{Lemma}

\newtheorem{theorem}{Theorem}
\newtheorem{corollary}{Corollary}

\newenvironment{ftheorem}
{\begin{mdframed}[hidealllines=true,backgroundcolor=mybox2 ,innerleftmargin=3pt,innerrightmargin=3pt,leftmargin=-3pt,rightmargin=-3pt]\begin{theorem}}
{\end{theorem}\end{mdframed}}

\definecolor{mygreen}{RGB}{0 139 69}
\definecolor{mybox2}{RGB}{230 230 250}
\definecolor{mybox}{RGB}{255 218 185}
\definecolor{myred}{RGB}{205 38 38}
\definecolor{mycyan}{cmyk}{.3,0,0,0}

\hypersetup{
	colorlinks=true,
	urlcolor=magenta,
}

\icmltitlerunning{Robustness and Accuracy Could Be Reconcilable by (Proper) Definition}

\begin{document}

\twocolumn[
\icmltitle{Robustness and Accuracy Could Be Reconcilable by (Proper) Definition}

\icmlsetsymbol{equal}{*}

\begin{icmlauthorlist}
\icmlauthor{Tianyu Pang}{to1,to2}
\icmlauthor{Min Lin}{to2}
\icmlauthor{Xiao Yang}{to1}
\icmlauthor{Jun Zhu}{to1}
\icmlauthor{Shuicheng Yan}{to2}
\end{icmlauthorlist}

\icmlaffiliation{to1}{Dept. of Comp. Sci. and Tech., Institute for AI, BNRist Center, THBI Lab, Tsinghua-Bosch Joint Center for ML, Tsinghua University.}
\icmlaffiliation{to2}{Sea AI Lab, Singapore}

\icmlcorrespondingauthor{Jun Zhu}{dcszj@tsinghua.edu.cn}
\icmlcorrespondingauthor{Shuicheng Yan}{yansc@sea.com}

\icmlkeywords{Machine Learning, ICML}

\vskip 0.3in
]

\printAffiliationsAndNotice{}  

\begin{abstract}
The trade-off between robustness and accuracy has been widely studied in the adversarial literature. Although still controversial, the prevailing view is that this trade-off is inherent, either empirically or theoretically. Thus, we dig for the origin of this trade-off in adversarial training and find that it may stem from the improperly defined robust error, which imposes an inductive bias of local invariance --- an overcorrection towards smoothness. Given this, we advocate employing local equivariance to describe the ideal behavior of a robust model, leading to a self-consistent robust error named SCORE. By definition, SCORE facilitates the reconciliation between robustness and accuracy, while still handling the worst-case uncertainty via robust optimization. By simply substituting KL divergence with variants of distance metrics, SCORE can be efficiently minimized. Empirically, our models achieve top-rank performance on RobustBench under AutoAttack. Besides, SCORE provides instructive insights for explaining the overfitting phenomenon and semantic input gradients observed on robust models.
\end{abstract}

\vspace{-0.55cm}
\section{Introduction}
\vspace{-0.075cm}
The trade-off between \emph{adversarial} robustness and \emph{clean} accuracy has been widely observed~\citep{schmidt2018adversarially,Su_2018_ECCV,zhang2019theoretically,wang2020once,wang2019improving}. 
On some simple cases, this trade-off is even shown to provably exist~\citep{tsipras2018robustness,nakkiran2019adversarial,raghunathan2020understanding,javanmard2020precise,yu2021understanding}. 
With that, many strategies have been proposed to alleviate this trade-off via, e.g., early-stopping~\citep{rice2020overfitting,zhang2020attacks}, instance reweighting~\citep{balaji2019instance,zhang2021geometryaware}, and exploiting extra data~\citep{alayrac2019labels,carmon2019unlabeled,hendrycks2019using}, to name a few.

Nevertheless, other findings show the opposite. \citet{stutz2019disentangling} and \citet{yang2020closer} argue that robustness and accuracy can both be achievable through manifold analyses or locally Lipschitz functions. \citet{rozsa2016accuracy} and \citet{gilmer2018adversarial} also reveal that better generalization helps robustness on both toy and large-scale datasets.

Given arguments on both sides, it is much debated whether the robustness-accuracy trade-off is intrinsically there.
But before considering its existence, let us first reach a consensus on how a robust model is supposed to behave --- the definition of robustness. 
A most popular one is made by \citet{madry2018towards}, who define robustness in terms of robust error formulated as a locally maximized loss function. This definition is widely adopted in adversarial training (AT)~\citep{shafahi2019adversarial,wong2020fast}.
Besides, there are also some other definitions. 
For example, \citet{Szegedy2013} define robustness as the minimal perturbation required to flip the predicted labels, which is applied in several adversarial attack studies~\citep{Moosavidezfooli2016,Carlini2016}.

We then try to describe how the robustness-accuracy trade-off stems from the previously defined robust error. Recall that in supervised learning, we have a joint data distribution $p_{d}(x,y)$, and a discriminative model $p_{\theta}(y|x)$ for the label $y$, which is conditional on the input $x$. In standard settings, we obtain an accurate model via minimizing the expected KL divergence between $p_{d}(y|x)$ and $p_{\theta}(y|x)$, i.e., the \textbf{standard error} $\mathbf{R}_{\textrm{Standard}}(\theta)$ (see Eq.~(\ref{eq11})) w.r.t. the parameters $\theta$. The optimal solution $\theta^{*}$ satisfies $p_{\theta^{*}}(y|x)=p_{d}(y|x)$.

\begin{figure*}[t]
\begin{center}
\vspace{-0.cm}
\includegraphics[width=2.00\columnwidth]{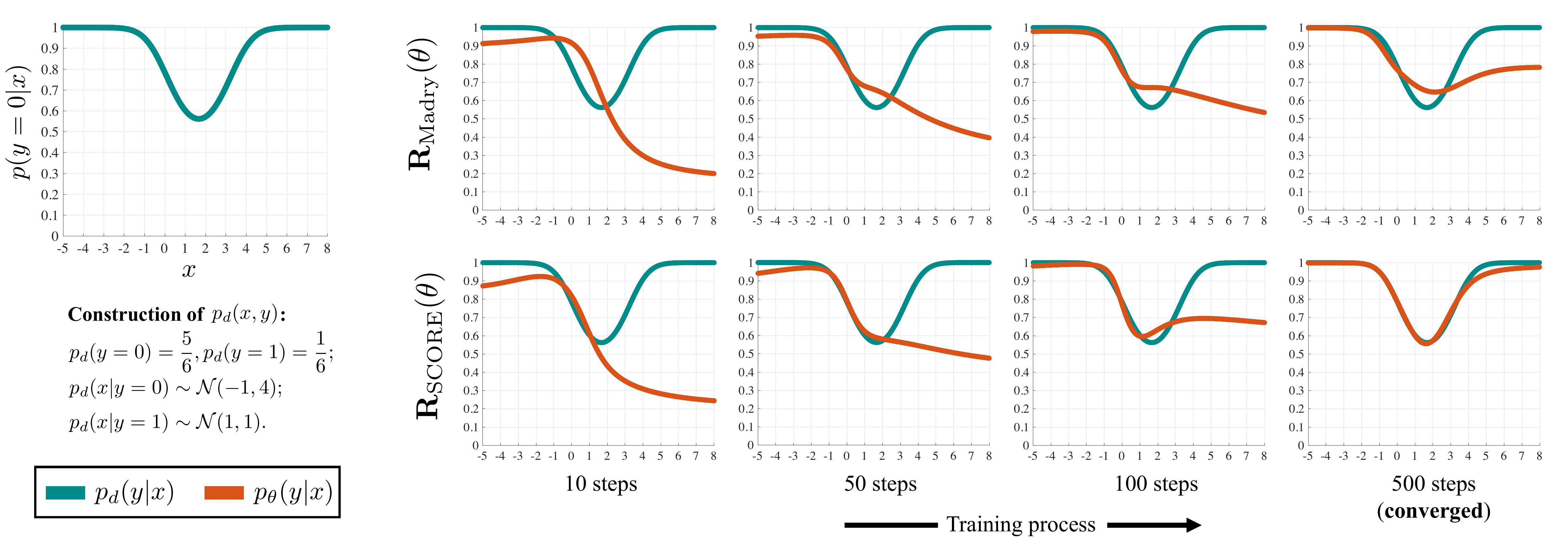}
\vspace{-0.15cm}
\caption{Toy demo on training $p_{\theta}(y|x)$ with $\mathbf{R}_{\textrm{Madry}}(\theta)$ and $\mathbf{R}_{\textrm{SCORE}}(\theta)$, respectively. We consider 1-d binary classification where $x\in \mathbb{R}$ and $y\in\{0,1\}$, and the perturbed set $B(x)=\{x'\big||x'-x|\leq 1\}$. Totally 60,000 training input-label pairs are sampled from the data distribution $p_{d}(x,y)$, as detailed in the left panel. The model $p_{\theta}(y|x)$ is a shallow MLP with two hidden layers and a sigmoid output. We use Adam optimizer and train for 500 steps to ensure convergence. The small gap between the converged $p_{\theta}(y|x)$ and $p_{d}(y|x)$ in the SCORE case is caused by limited model capacity. Note that in toy demo we assume $p_{d}(y|x)$ is accessible, which is not true in practice.}
\label{fig:1}
\end{center}
\vspace{0.15cm}
\end{figure*}

In adversarial settings, \citet{madry2018towards} propose to minimize the \textbf{robust error} $\mathbf{R}_{\textrm{Madry}}(\theta)$ (see Eq.~(\ref{eq22})) for training reliable models. 
Compared to the standard error, the robust error contains an inner maximization problem, finding the point $x'\in B(x)$ that maximizes the KL divergence between $p_{d}(y|x)$ and $p_{\theta}(y|x')$, where $B(x)$ is a set of allowed points around $x$. 
Minimizing the robust error w.r.t. $\theta$ imposes an inductive bias towards local invariance: for $\forall x'\in B(x)$, $p_{\theta}(y|{x'})$ is encouraged to be equal to $p_{d}(y|{x})$.
This locally-invariant bias makes the learned model tend to be over-smoothed, as observed in previous works~\citep{stutz2019confidence,chen2020robust}.
Hence, generally $p_{\theta^{*}}(y|x)\neq p_{d}(y|x)$ for the optimal $\theta^{*}$ of $\min_{\theta}\mathbf{R}_{\textrm{Madry}}(\theta)$. In Fig.~\ref{fig:1}, we observe that $p_{\theta^{*}}(y|x)$ does not converge to $p_{d}(y|{x})$ even on toy examples. 
The ending that $p_{d}(y|{x})$ is not an optimally robust model w.r.t. $p_{d}(y|{x})$ itself seems counter-intuitive, revealing that $\mathbf{R}_{\textrm{Madry}}(\theta)$ inherently does not support the reconciliation between robustness and accuracy.

Formally resolving the above inconsistency motivates us to properly redefine the robust error. To be specific, we substitute the inductive bias of local invariance with local equivariance, i.e., $\forall x'\in B(x)$, $p_{\theta}(y|x')$ is encouraged to point-wisely stick to $p_{d}(y|x')$, leading to the definition of \textbf{Self-COnsistent Robust Error (SCORE)} as $\mathbf{R}_{\textrm{SCORE}}(\theta)$ (see Eq.~(\ref{eq33})). 
Compared to the robust error, SCORE finds the point $x'$ that maximizes the KL divergence between $p_{d}(y|x')$ and $p_{\theta}(y|x')$ in its inner problem. SCORE aligns the optimal solution $p_{\theta^{*}}(y|x)$ with $p_{d}(y|x)$ (i.e., self-consistency, as shown in Fig.~\ref{fig:1}), while keeping the paradigm of robust optimization~\citep{wald1945statistical} in the finite-sample cases, as demonstrated in Fig.~\ref{fig:2}. More details can be found in Sec.~\ref{sec2}.

\textbf{How to optimize SCORE?} Note that we only have closed-form access to $p_{d}(y|x)$ in toy cases. In practice, minimizing SCORE by the off-the-shelf first-order optimizers (e.g., SGD or Adam) requires estimating $\nabla_{x}\log p_{d}(y|x)$ (see Appendix~\ref{secB1}). This task can be decomposed into score matching tasks~\citep{vincent2011connection} estimating the data scores $\nabla_{x}\log p_{d}(x)$ and $\nabla_{x}\log p_{d}(x|y)$, respectively. However, our initial experiments show that the estimated data scores are of high variance, making it non-trivial to well adopt them in the discriminative learning process. Fortunately, as described in Sec.~\ref{sec3}, we find that by replacing KL divergence with any metric $\mathcal{D}$ satisfying the distance axioms (symmetry, triangle inequality), we can derive upper and lower bounds for distance-based SCORE $\mathbf{R}_{\textrm{SCORE}}^{\mathcal{D}}(\theta)$, and minimize it without knowing $\nabla_{x}\log p_{d}(y|x)$.

In Sec.~\ref{sec4}, we bridge the gap between distance-based SCORE $\mathbf{R}_{\textrm{SCORE}}^{\mathcal{D}}(\theta)$ and previously used KL-based objectives like $\mathbf{R}_{\textrm{Madry}}(\theta)$ via Pinsker's inequality. This connection inspires us to propose instructive explanations for some well-known phenomena of robust models, e.g., overfitting~\citep{rice2020overfitting} and semantic gradients~\citep{ilyas2019adversarial}.

In Sec.~\ref{sec6}, we validate the effectiveness of replacing KL divergence with distance-based metrics (and their variants), developed from the analyses of SCORE. We improve the state-of-the-art AT methods under AutoAttack~\citep{croce2020reliable}, and achieve top-rank performance with 1M DDPM generated data on the leader boards of CIFAR-10 and CIFAR-100 on RobustBench~\citep{croce2020robustbench}.

\section{Self-Consistent Robust Error}
\label{sec2}
According to \citet{madry2018towards}, robustness connects to a certain definition of robust error, which means a more robust model is a one that achieves lower robust error.\footnote{Here, robustness refers to its differentiable surrogate form. In Sec.~\ref{sec51} we will discuss its 0-1 definition used for evaluation.} In this section, we first revisit previous definitions of robustness, and then propose a self-consistent robust error.

\subsection{Preliminaries}
In supervised learning, a training set $\{(x^{i},y^{i})\}_{i=1}^{N}$ consists of $N$ i.i.d. input-label pairs $(x^{i},y^{i})$ sampled from the joint data distribution $p_{d}(x,y)$. Let $p_{\theta}(y|x)$ be a discriminative model parameterized by $\theta$. It is trained to match $p_{d}(y|x)$ by minimizing the standard error:
\begin{equation}
     \mathbf{R}_{\textrm{Standard}}(\theta)=\mathbb{E}_{ p_{d}(x)}\left[\textrm{KL}\left(p_{d}(y|x)\big\|p_{\theta}(y|x)\right)\right]\textrm{,}
     \label{eq11}
\end{equation}
where $\textrm{KL}(P\|Q)$ denotes the KL divergence between two distributions $P$ and $Q$. Since the data distribution $p_{d}$ is independent of $\theta$, minimizing $\mathbf{R}_{\textrm{Standard}}(\theta)$ w.r.t. $\theta$ is equivalent to minimizing the cross-entropy loss~\citep{friedman2001elements}, where the optimal solution is $p_{\theta^*}(y|x)=p_{d}(y|x)$.

\subsection{Definitions of Robustness}
There are several definitions of adversarial robustness in literature. For example, the seminal work of \citet{Szegedy2013} defines robustness as the minimal perturbation required to flip the predicted labels:
\begin{equation}
    \mathbf{R}(x,\theta)=\min_{\delta}\|\delta\|\textrm{, s.t. }\mathcal{Y}_{\theta}(x+\delta)\neq\mathcal{Y}_{\theta}(x)\textrm{,}
    \label{eq35}
\end{equation}
where $\mathcal{Y}_{\theta}(x)=\argmax_{y} p_{\theta}(y|x)$. 
Following this definition, several adversarial attacks are developed, aiming to find the successful evasions with minimal norms~\citep{Carlini2016,brendel2019accurate,rony2019decoupling,pintor2021fast}; margin-based defenses are also proposed~\citep{tsuzuku2018lipschitz,pang2018max,Ding2020MMA}. 
But solving the problem in Eq.~(\ref{eq35}) is computationally expensive, making it an intractable objective for end-to-end training. Thus, it is less discussed in the defense literature.

In contrast, \citet{madry2018towards} propose PGD-AT and define robustness in terms of the robust error that is well compatible with supervised learning, formulated as
\begin{equation}
    \!\mathbf{R}_{\textrm{Madry}}(\theta)\!=\!\mathbb{E}_{p_{d}(x)}\!\left[\max_{{\color{red}x'}\in B(x)}\!\!\textrm{KL}\left(p_{d}(y|x)\big\|p_{\theta}(y|{\color{red}x'})\right)\right]\textrm{.}
     \label{eq22}
\end{equation}
This definition is the most commonly used one, especially for adversarial training (AT)~\citep{kannan2018adversarial,wong2020fast,shafahi2019adversarial,rice2020overfitting}. Notice that the original definition in \citet{madry2018towards} is formulated by the cross-entropy loss, while Eq.~(\ref{eq22}) can be regarded as its expected version using KL divergence. In Appendix~\ref{Rmardy}, we show that these two versions are equivalent under first-order optimization.

\begin{figure}[t]
\begin{center}
\vspace{-0.cm}
\includegraphics[width=1.00\columnwidth]{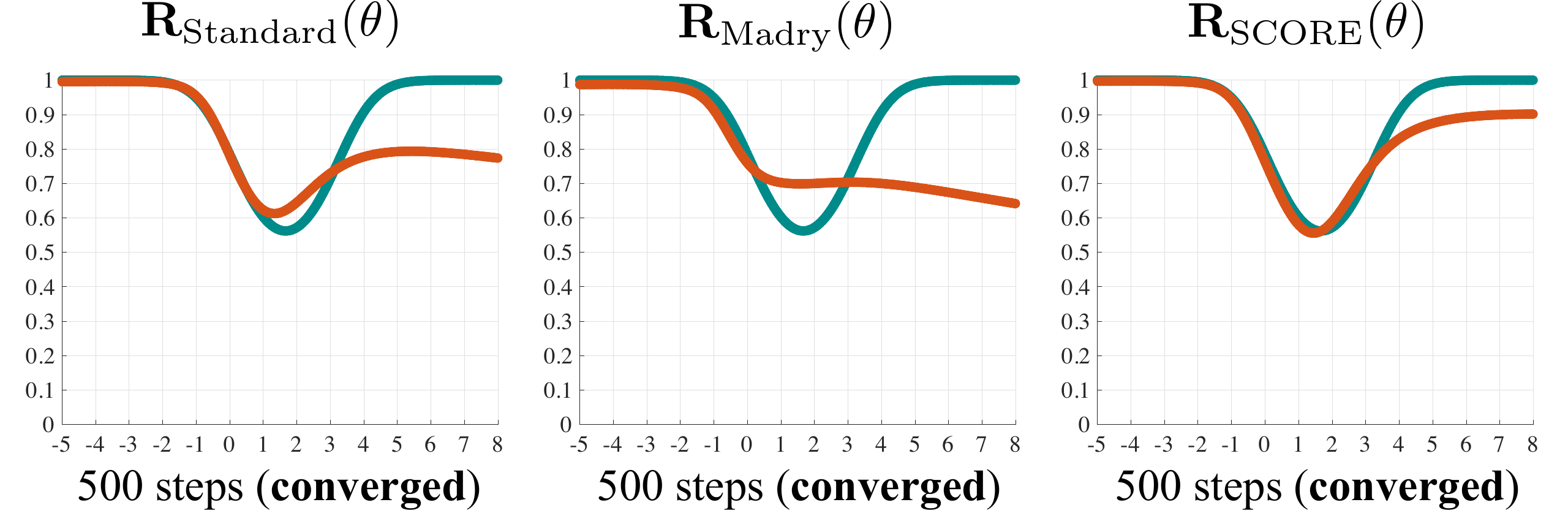}
\vspace{-0.7cm}
\caption{Note that Fig.~\ref{fig:1} samples 60,000 training pairs, which suffice to well approximate the expectation $\mathbb{E}_{p_{d}(x)}$ in a 1-d toy problem. In contrast, here we only use 6 training pairs to mimic the finite-sample cases encountered in practice. We plot the converged states after 500 training steps for $\mathbf{R}_{\textrm{Standard}}(\theta)$, $\mathbf{R}_{\textrm{Madry}}(\theta)$, and $\mathbf{R}_{\textrm{SCORE}}(\theta)$, respectively.
As can be seen, training with $\mathbf{R}_{\textrm{SCORE}}(\theta)$ inherits the benefits of robust optimization with a self-consistent optimality, which improves the sample efficiency of learning.}
\label{fig:2}
\end{center}
\vspace{-0.15cm}
\end{figure}

Note that for $\forall x'\in B(x)$, $\mathbf{R}_{\textrm{Madry}}(\theta)$ encourages $p_{\theta}(y|{x'})$ to be locally invariant and equal to $p_{d}(y|{x})$. 
As demonstrated in Fig.~\ref{fig:1}, generally there is $p_{\theta^*}(y|x)\neq p_{d}(y|x)$ when minimizing $\mathbf{R}_{\textrm{Madry}}(\theta)$. It is non-trivial to derive a closed-form solution for $p_{\theta^*}(y|x)$ even if we set $p_{d}(y|x)$ to be simple toy distributions.
Previous works observe that the model learned by minimizing $\mathbf{R}_{\textrm{Madry}}(\theta)$ and its variants like TRADES~\citep{zhang2019theoretically} will lead to over-smoothed decision landscapes~\citep{stutz2019confidence,chen2020robust}.

\subsection{A Self-Consistent Robust Error}
If we admit $\mathbf{R}_{\textrm{Madry}}(\theta)$ as the proper definition of robust error, we would arrive at a paradox that $p_{d}(y|x)$ is not an optimally robust model w.r.t. $p_{d}(y|x)$ itself, which contradicts the basic preconditions of supervised learning.

Thus, we suppose the misalignment between $p_{\theta^*}(y|x)$ and $p_{d}(y|x)$ when minimizing $\mathbf{R}_{\textrm{Madry}}(\theta)$ is one of the essential reasons for the trade-off between robustness and accuracy. 
To eliminate this trade-off, we slightly modify the definition of robust error and propose the \textbf{Self-COnsistent Robust Error (SCORE)}, which is formulated as
\begin{equation}
    \!\!\!\mathbf{R}_{\textrm{SCORE}}(\theta)\!=\!\mathbb{E}_{p_{d}(x)}\!\left[\max_{{\color{red}x'}\in B(x)}\!\!\textrm{KL}\left(p_{d}(y|{\color{red}x'})\big\|p_{\theta}(y|{\color{red}x'})\right)\right]\!\textrm{.}\!
     \label{eq33}
\end{equation}
Upon the inductive bias of local invariance imposed by $\mathbf{R}_{\textrm{Madry}}(\theta)$, SCORE makes amendment with local equivariance, allowing $p_{\theta}(y|x')$ to point-wisely match $p_{d}(y|x')$ for any $x'\in B(x)$. 
The self-consistency is thus achieved, or namely the optimal solution for minimizing $\mathbf{R}_{\textrm{SCORE}}(\theta)$ w.r.t. $\theta$ is $p_{\theta^*}(y|x)=p_{d}(y|x)$. In finite-sample cases, $\mathbf{R}_{\textrm{SCORE}}(\theta)$ preserves the paradigm of robust optimization~\citep{wald1945statistical} to cover the worst-case uncertainty, extracting more information from the samples in hand as seen in Fig.~\ref{fig:2}. The effect of promoting sample efficiency via robust optimization is also observed on large-scale tasks~\citep{xie2019adversarial}.

\vspace{-0.13cm}
\section{How to Practically Optimize SCORE?}
\vspace{-0.02cm}
\label{sec3}
While SCORE seems a promising objective, particularly for adversarial training, it is intractable to directly optimize $\mathbf{R}_{\textrm{SCORE}}(\theta)$ in practice. That is, minimizing $\mathbf{R}_{\textrm{SCORE}}(\theta)$ with existing first-order optimizers (e.g., SGD or Adam) requires closed-form access to $\nabla_{x}\log p_{d}(y|x)$ (see Appendix~\ref{secB1}). Although we can obtain training samples from $p_{d}(x,y)$, we cannot differentiate through the real data distribution. To this end, generative methods like score matching may be applied to estimate $\nabla_{x}\log p_{d}(y|x)$~\citep{vincent2011connection}.

Nevertheless, through initial experiments we find that the estimated data scores are of high variance, and as a result, it is non-trivial to adopt them in the discriminative learning process. In this section, we elaborate on how to subtly avoid the need of directly optimizing $\mathbf{R}_{\textrm{SCORE}}(\theta)$ via resorting to distance metrics. Proofs of Theorems are in Appendix~\ref{proofs}.

\subsection{Substituting KL Divergence with Distance Metrics}
The KL divergence is \emph{not} a distance metric, since it is asymmetric and does not satisfy the triangle inequality~\citep{treves2016topological}. 
In contrast, a distance metric $\mathcal{D}(\cdot\|\cdot)$ satisfies the three axioms of identity of indiscernibles, symmetry and the triangle inequality: $ \mathcal{D}(A\|B)\leq \mathcal{D}(A\|C) + \mathcal{D}(C\|B)$. Typical examples include $\ell_{p}$-distances for $p\geq 1$, where $\|A-B\|_{p}\leq\|A-C\|_{p}+\|C-B\|_{p}$. By substituting KL divergence with a certain distance metric $\mathcal{D}(\cdot\|\cdot)$, we denote
\begin{equation}
    \!\mathbf{R}_{\textrm{Madry}}^{\mathcal{D}}(\theta)\!=\!\mathbb{E}_{p_{d}(x)}\!\left[\max_{{\color{red}x'}\in B(x)}\!\!\mathcal{D}\left(p_{d}(y|x)\big\|p_{\theta}(y|{\color{red}x'})\right)\right]\textrm{;}
     \label{eq32}
\end{equation}
\begin{equation}
    \!\!\!\mathbf{R}_{\textrm{SCORE}}^{\mathcal{D}}(\theta)\!=\!\mathbb{E}_{p_{d}(x)}\!\left[\max_{{\color{red}x'}\in B(x)}\!\!\mathcal{D}\left(p_{d}(y|{\color{red}x'})\big\|p_{\theta}(y|{\color{red}x'})\right)\right]\!\textrm{.}\!
     \label{eq34}
\end{equation}
Intriguingly, now we can derive upper and lower bounds for $\mathbf{R}_{\textrm{SCORE}}^{\mathcal{D}}(\theta)$ using $\mathbf{R}_{\textrm{Madry}}^{\mathcal{D}}(\theta)$, as stated below:
\begin{ftheorem}
\label{theorem1}
(Bounding the SCORE objective) For any distance metric $\mathcal{D}(\cdot||\cdot)$, there are lower and upper bounds that
\begin{equation*}
\begin{split}
    &\lvert{\mathbf{R}_{\textup{Madry}}^{\mathcal{D}}(\theta) - C^{\mathcal{D}}}\rvert\leq \mathbf{R}_{\textup{SCORE}}^{\mathcal{D}}(\theta) \leq \mathbf{R}_{\textup{Madry}}^{\mathcal{D}}(\theta) + C^{\mathcal{D}}\textrm{,}\\
    &\textrm{where }C^{\mathcal{D}} = \mathbb{E}_{p_{d}(x)}\left[\max_{{\color{red}x'}\in B(x)}\mathcal{D}\left(p_{d}(y|x)\big\|p_{d}(y|{\color{red}x'})\right)\right]
    \end{split}
\end{equation*}
is a constant independent of $\theta$. The lower bound becomes tight when $\mathbf{R}_{\textup{SCORE}}^{\mathcal{D}}(\theta)=0$, and we have $\mathbf{R}_{\textup{Madry}}^{\mathcal{D}}(\theta) = C^{\mathcal{D}}$.
\end{ftheorem}
\textbf{What is the constant $C^{\mathcal{D}}$?} The constant $C^{\mathcal{D}}$ indicates the intrinsic smoothness of the data distribution, i.e., how much $p_{d}(y|x)$ has changed in the neighborhood set $B(x)$.
Though we cannot precisely compute $C^{\mathcal{D}}$ in practice, more complex datasets are expected to have larger values of $C^{\mathcal{D}}$.

\textbf{Upper bound.} $\mathbf{R}_{\textup{SCORE}}^{\mathcal{D}}(\theta) \leq \mathbf{R}_{\textup{Madry}}^{\mathcal{D}}(\theta) + C^{\mathcal{D}}$ guarantees that we can relax the goal from minimizing $\mathbf{R}_{\textup{SCORE}}^{\mathcal{D}}(\theta)$ to minimizing $\mathbf{R}_{\textup{Madry}}^{\mathcal{D}}(\theta)$, without the need to estimate data scores.
A similar trick of optimizing upper bounds is widely applied in variational learning~\citep{kingma2013auto}.

\textbf{Lower bound.} $\lvert{\mathbf{R}_{\textup{Madry}}^{\mathcal{D}}(\theta) - C^{\mathcal{D}}}\rvert\leq \mathbf{R}_{\textup{SCORE}}^{\mathcal{D}}(\theta)$ tells us that: (\textbf{\romannumeral 1}) $\mathbf{R}_{\textup{Madry}}^{\mathcal{D}}(\theta)=C^{\mathcal{D}}$ is a necessary condition for $\mathbf{R}_{\textup{SCORE}}^{\mathcal{D}}(\theta)=0$; (\textbf{\romannumeral 2}) overly minimizing $\mathbf{R}_{\textup{Madry}}^{\mathcal{D}}(\theta)$ to approach zero makes $\mathbf{R}_{\textup{SCORE}}^{\mathcal{D}}(\theta)$ tend to increase or overfit.
In Sec.~\ref{sec41}, we will show that a similar conclusion holds for the original KL-based $\mathbf{R}_{\textup{Madry}}(\theta)$, which closely connects to the overfitting phenomenon observed in~\citet{rice2020overfitting}.

\begin{figure}[t]
\begin{center}
\vspace{-0.cm}
\includegraphics[width=1.00\columnwidth]{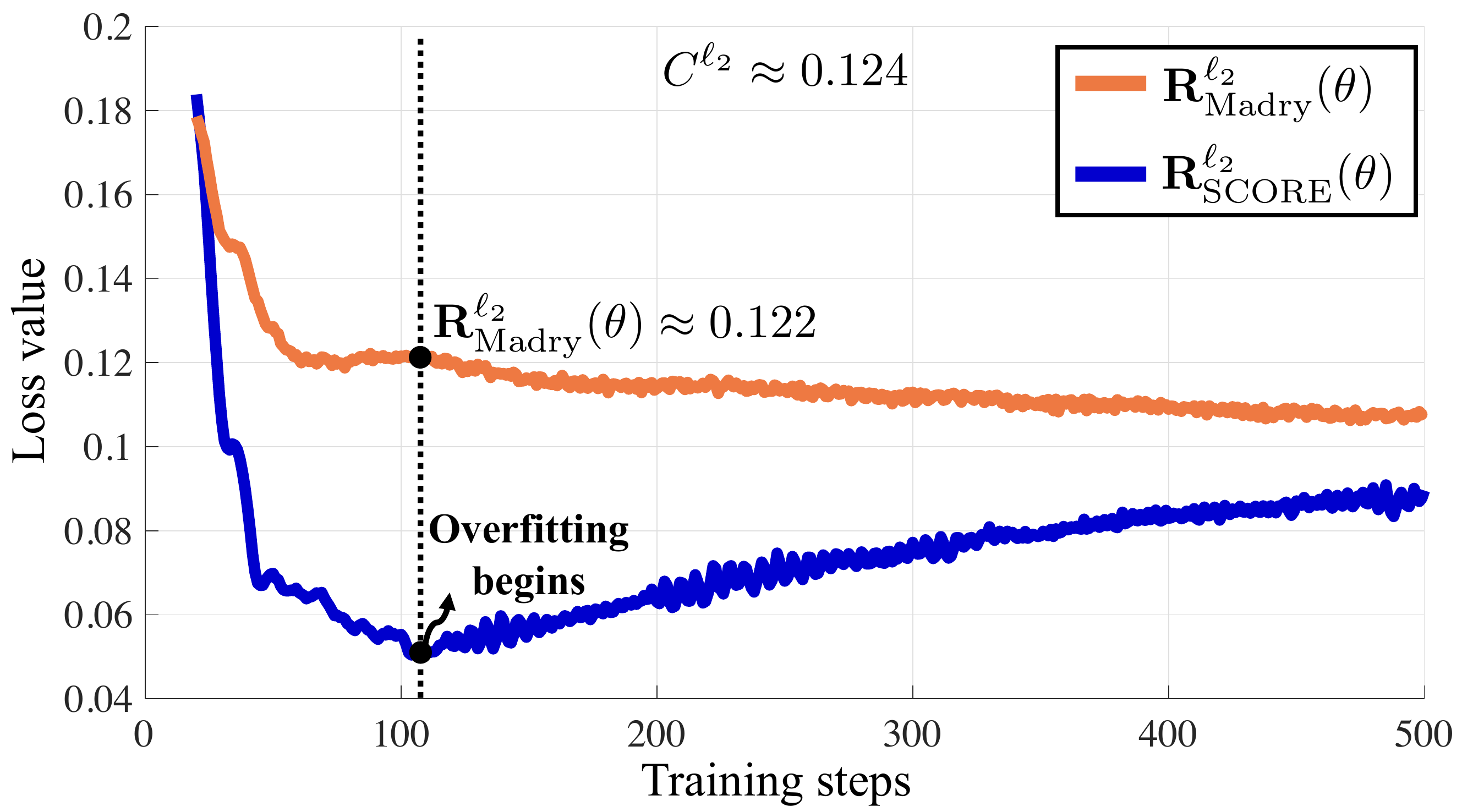}
\vspace{-0.7cm}
\caption{The overfitting phenomenon encountered when minimizing $\mathbf{R}_{\textup{Madry}}^{\mathcal{D}}(\theta)$. We use the same toy data distribution as in Fig.~\ref{fig:1}, and set $\mathcal{D}$ to be $\ell_{2}$-distance. As annotated, when $\mathbf{R}_{\textup{Madry}}^{\ell_{2}}(\theta)\approx C^{\ell_{2}}$ ($\approx 0.12$ in this example), the overfitting phenomenon happens, i.e., $\mathbf{R}_{\textup{SCORE}}^{\ell_{2}}(\theta)$ begins to increase while $\mathbf{R}_{\textup{Madry}}^{\ell_{2}}(\theta)$ still decreases.}
\label{fig:4}
\end{center}
\vspace{-0.15cm}
\end{figure}

\textbf{A toy demo.} In Fig.~\ref{fig:4}, we leverage a toy example to further explain the above properties claimed in Theorem~\ref{theorem1}. 
During training, we minimize $\mathbf{R}_{\textup{Madry}}^{\ell_{2}}(\theta)$, and the constant $C^{\ell_{2}}$ can be computed in closed-form. 
As can be seen, in the initial phase of training (first $\sim100$ training steps), the upper bound works, and $\mathbf{R}_{\textup{SCORE}}^{\ell_{2}}(\theta)$ is effectively minimized.
Then after $\mathbf{R}_{\textup{Madry}}^{\ell_{2}}(\theta)$ is minimized to be less than $C^{\ell_{2}}$, the lower bound intervenes and overfitting happens.
Intuitively, when $\mathbf{R}_{\textup{Madry}}^{\mathcal{D}}(\theta)<C^{\mathcal{D}}$, the learned model $p_{\theta}(y|x)$ is considered as over-smoothed compared to the oracle $p_{d}(y|x)$.

\subsection{Monotonically Increasing Convex Variants}
In the above, we show that substituting KL divergence with any distance metric $\mathcal{D}$ can induce favorable bounds, which enable us to optimize SCORE efficiently and forebode the overfitting phenomenon. 
However, the KL divergence involves logarithm function like $\log p_{\theta}(y|x)$, whose gradients may focus on unlearned examples (with small values of $p_{\theta}(y|x)$). 
In contrast, a distance metric is sublinear and does not work well in practice, as empirically shown in Table~\ref{table3}. 
Thus, we generalize Theorem~\ref{theorem1} to monotonically increasing convex variants of $\mathcal{D}$, formalized as below:
\begin{ftheorem}
\label{theorem3}
(Variants of $\mathcal{D}$) For any distance metric $\mathcal{D}(\cdot||\cdot)$ and a monotonically increasing convex function $\phi(\cdot)$,
\begin{equation*}
    \lvert{\mathbf{R}_{\textup{SCORE}}^{\mathcal{D}}(\theta)-C^{\mathcal{D}}\rvert} \leq \phi^{-1}\left(\mathbf{R}_{\textup{Madry}}^{\phi\circ\mathcal{D}}(\theta)\right)\textrm{,}
\end{equation*}
where $\phi^{-1}(\cdot)$ is the inverse function, and the superscript $\phi\circ\mathcal{D}$ means using the composition $\phi\circ\mathcal{D}$ in Eq.~(\ref{eq32}).
\end{ftheorem}
\textbf{Remark.} Theorem~\ref{theorem3} allows us to construct upper bounds of $\mathbf{R}_{\textup{SCORE}}^{\mathcal{D}}(\theta)$ utilizing more general variants of distance, e.g., squared error (SE) $\|P-Q\|_{2}^{2}$ or JS divergence $\textrm{JS}(P\|Q)$, based on the fact that $\sqrt{\textrm{JS}(P\|Q)}$ is a distance metric~\citep{endres2003new}. 
These variants work much better than their distance counterparts, as seen in Table~\ref{table3}.

In our experiments, we take SE as an example derived from our analyses, and verify that substituting KL divergence with SE in training objectives improves the performance of state-of-the-art AT methods, as will be detailed in Sec.~\ref{sec6}.

\subsection{Equivalent Relation Induced by Distance Metrics}
Besides PGD-AT~\citep{madry2018towards}, another typical AT method is TRADES~\citep{zhang2019theoretically}. Given any distance metric $\mathcal{D}$, $\mathbf{R}_{\textup{Madry}}^{\mathcal{D}}(\theta)$ and $\mathbf{R}_{\textup{TRADES}}^{\mathcal{D}}(\theta;\beta)$ (defined in Eq.~(\ref{eq18})) are equivalent in the parameter space, i.e., induce the same topology of loss landscapes~\citep{conrad2018equivalence}:
\begin{ftheorem}
\label{theorem2}
(Equivalent Relation) For any distance metric $\mathcal{D}(\cdot||\cdot)$ and a given hyperparameter $\beta\geq 1$, there is
\begin{equation*}
    \mathbf{R}_{\textup{Madry}}^{\mathcal{D}}(\theta)\leq\mathbf{R}_{\textup{TRADES}}^{\mathcal{D}}(\theta;\beta)\leq \left(1+2\beta\right)\cdot\mathbf{R}_{\textup{Madry}}^{\mathcal{D}}(\theta)\textrm{,}
\end{equation*}
which holds for any model parameters $\theta$.
\end{ftheorem}
Therefore, the conclusions that hold for $\mathbf{R}_{\textup{Madry}}^{\mathcal{D}}(\theta)$ similarly hold for $\mathbf{R}_{\textup{TRADES}}^{\mathcal{D}}(\theta;\beta)$, as justified in our experiments.

\section{New Insights Brought by SCORE}
\label{sec4}
Although KL divergence is not a distance metric, Pinsker's inequality~\citep{csiszar2011information} claims that
\begin{equation}
    \frac{1}{2}\|P-Q\|_{1}^{2}\leq{\mathrm{KL}(P||Q)}
\end{equation}
holds for two distributions $P$ and $Q$. This connects KL divergence with distance metrics, leading to some interesting views for explaining previous observations on robust models, like overfitting and semantic input gradients. We would not say that our explanations are conclusive; instead, we aim to provide insights to explore these phenomena further.

\subsection{Overfitting and Early-Stopping}
\label{sec41}
\citet{rice2020overfitting} observe that overfitting happens in the process of adversarial training, and simply using early-stopping can reduce the robust generalization gap.
Previous works attribute the overfitting phenomenon to activation functions~\citep{singla2021low}, perturbation underfitting~\citep{li2020overfitting} or hard instances~\citep{liu2021impact}.
In contrast, we find a more straightforward explanation from the view of SCORE. 
Specifically, according to Theorem~\ref{theorem3} and Pinsker's inequality, there is
\begin{corollary}
\label{corollary1}
Let $\ell_{1}$ refers to $\ell_{1}$-distance metric. There is
\begin{equation*}
    \lvert{\mathbf{R}_{\textup{SCORE}}^{\ell_{1}}(\theta)-C^{\ell_{1}}\rvert} \leq \sqrt{2\cdot\mathbf{R}_{\textup{Madry}}(\theta)}\textrm{.}
\end{equation*}
\end{corollary}
The square-root robust error $\sqrt{2\cdot\mathbf{R}_{\textup{Madry}}(\theta)}$ acts as an upper bound for the SCORE $\mathbf{R}_{\textup{SCORE}}^{\ell_{1}}(\theta)$ in the initial phase of training (i.e., when $\mathbf{R}_{\textup{SCORE}}^{\ell_{1}}(\theta)>C^{\ell_{1}}$), which coincides with the observation that accuracy and robustness can both increase before overfitting happens~\citep{rice2020overfitting}. As can be seen, the necessary condition for $\mathbf{R}_{\textup{SCORE}}^{\ell_{1}}(\theta)=0$ is
\begin{equation}
    C^{\ell_{1}}\leq \sqrt{2\cdot\mathbf{R}_{\textup{Madry}}(\theta)} \Longrightarrow \mathbf{R}_{\textup{Madry}}(\theta)\geq\frac{\left(C^{\ell_{1}}\right)^{2}}{2}\textrm{,}
\end{equation}
which implies $\mathbf{R}_{\textup{Madry}}(\theta)$ should be early-stopped at least before ${\left(C^{\ell_{1}}\right)^{2}}/{2}$ to avoid the overfitting phenomenon. 
In Appendix~\ref{secD2}, we show that the overfitting actually happens even earlier at $\mathbf{R}_{\textup{Madry}}(\theta)\approx C^{\textrm{KL}}$ (demo in Fig.~\ref{appendixfig:2} (a)).

\begin{figure}[t]
\begin{center}
\vspace{-0.cm}
\includegraphics[width=1.00\columnwidth]{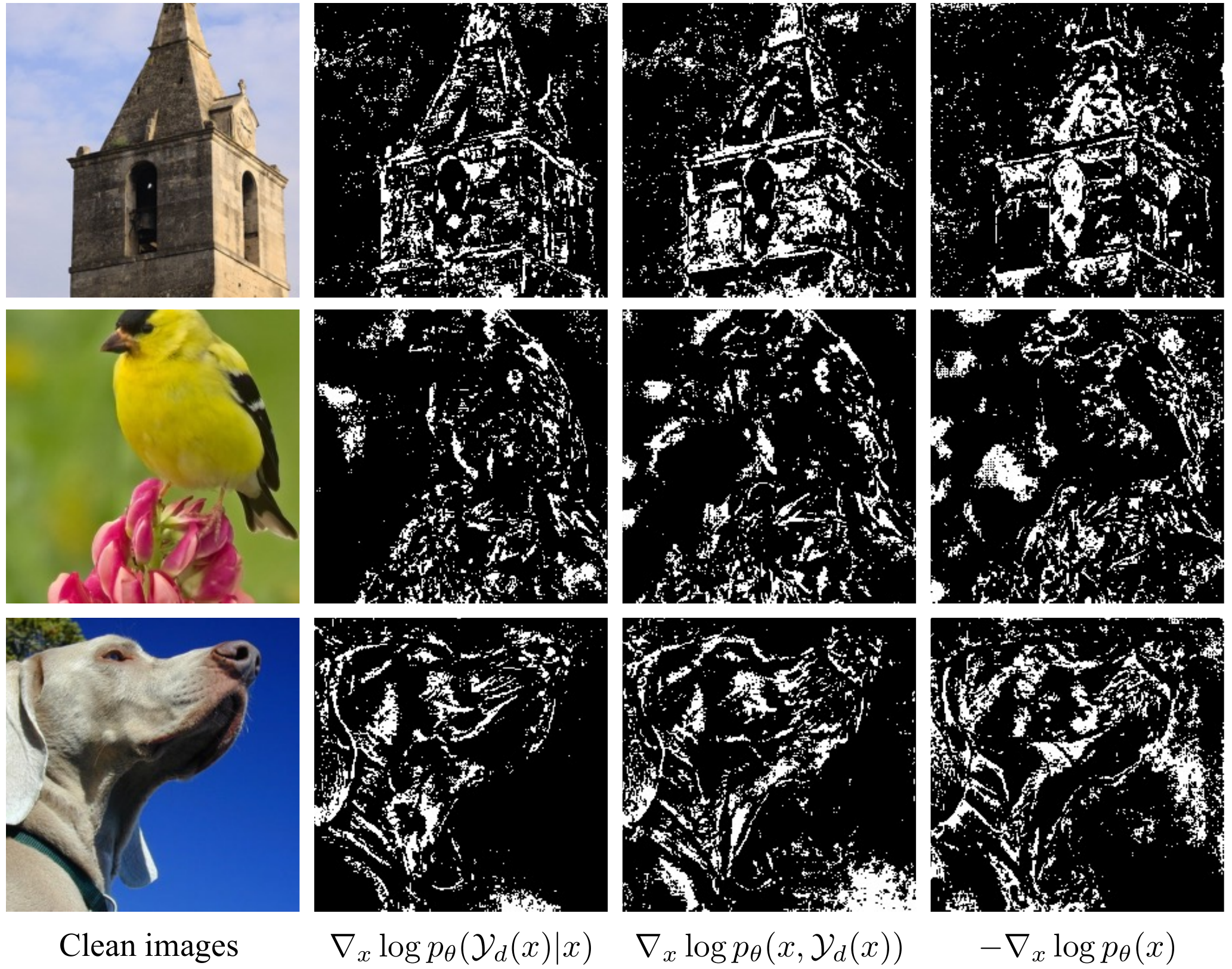}
\vspace{-0.7cm}
\caption{Visualization of semantic gradients. We adopt a ResNet-50 model adversarially trained by FreeAT on ImageNet. Here $\nabla_{x}\log p_{\theta}(x,\mathcal{Y}_{d}(x))$ and $-\nabla_{x}\log p_{\theta}(x)$ are constructed according to \citet{Grathwohl2020Your} (details in Appendix~\ref{VSG}), and we set $\mathcal{Y}_{d}(x)$ to be the test label of each clean image. We add up the partial derivatives of three RGB channels for each pixel position, and sort out the top 10\% pixel positions with large values of total derivatives (i.e., those affect the objectives the most) in the plots.}
\label{fig:3}
\end{center}
\vspace{-0.1cm}
\end{figure}

\subsection{Semantic Gradients: Adversarial Training}
\label{sec42}
Previous studies observe that the input gradients $\nabla_{x}p_{\theta}(y|x)$ of the adversarially trained models exhibit semantic or perceptually-aligned characteristics~\citep{tao2018attacks,ilyas2019adversarial,santurkar2019image,etmann2019connection,chan2019jacobian}. 
Recall that minimizing $\mathbf{R}_{\textup{Madry}}(\theta)$ effectively minimizes the gap between $\mathbf{R}_{\textup{SCORE}}^{\ell_{1}}(\theta)$ and $C^{\ell_{1}}$. Considering this, in the $\ell_{p}$-norm threat model of $B(x)=\{x'\big|\|x'-x\|_{p}\leq \epsilon\}$, we take the first-order expansion similar as in \citet{simon2019first}, and obtain 
\begin{ftheorem}
\label{theorem4}
(Gradient Alignment) Assume that in training, there is $p_{d}(y|x)>p_{\theta}(y|x)$ for $y=\mathcal{Y}_{d}(x)$; otherwise $p_{d}(y|x)<p_{\theta}(y|x)$. Then under first-order expansion,
\begin{equation*}
\begin{split}
    &\mathbf{R}_{\textup{SCORE}}^{\ell_{1}}(\theta) = \mathbf{R}_{\textup{Standard}}^{\ell_{1}}(\theta) + \\
    2\epsilon\!\cdot\!\mathbb{E}_{p_{d}(x)}\!&\left[\left\|\nabla_{x}p_{d}(\mathcal{Y}_{d}(x)|x)\!-\!\nabla_{x}p_{\theta}(\mathcal{Y}_{d}(x)|x)\right\|_{q}\right]\!+\!o(\epsilon)\textrm{,}
\end{split}
\end{equation*}
where $\mathbf{R}_{\textup{Standard}}^{\ell_{1}}(\theta)$ is the standard error using $\ell_{1}$-distance, $\mathcal{Y}_{d}(x)=\argmax_{y}p_{d}(y|x)$, and $\|\cdot\|_{q}$ is the dual of $\|\cdot\|_{p}$.
\end{ftheorem}
The assumption in Theorem~\ref{theorem4} is trivially true if we substitute $p_{d}(y|x)$ with a one-hot vector, as in supervised learning. Then, minimizing $\mathbf{R}_{\textup{Madry}}(\theta)$ (and thus $\mathbf{R}_{\textup{SCORE}}^{\ell_{1}}(\theta)$) encourages gradient alignment between $\nabla_{x}p_{d}(\mathcal{Y}_{d}(x)|x)$ and $\nabla_{x}p_{\theta}(\mathcal{Y}_{d}(x)|x)$ (demo in Fig.~\ref{appendixfig:2} (b)). This is analogous to score matching, but is not exactly the same since $p(y|x)$ is not normalized w.r.t. $x$, i.e., $\int_{x}p(y|x) dx\neq 1$. Still, adversarial training enjoys generative learning patterns.

In Fig.~\ref{fig:3}, we use a ResNet-50 trained by FreeAT~\citep{shafahi2019adversarial} on ImageNet, and visualize its input gradients $\nabla_{x}\log p_{\theta}(\mathcal{Y}_{d}(x)|x)$. 
We also plot $\nabla_{x}\log p_{\theta}(x, \mathcal{Y}_{d}(x))$ and $\nabla_{x}\log p_{\theta}(x)$ via regarding the classifier as an implicit EBM~\citep{Grathwohl2020Your}.
We highlight prominent pixels with large derivatives. 
As seen, adversarially trained models catch shape-based features, which are more generative compared to texture-based ones~\citep{geirhos2018imagenettrained}.

\subsection{Semantic Gradients: Randomized Smoothing}
\label{sec43}
Randomized smoothing is a promising method towards scalable certified defenses~\citep{cohen2019certified}. The smoothed model is trained by Gaussian augmentation and returns an ensemble prediction. Inspired by Theorem~\ref{theorem4}, we discover a similar gradient alignment objective in randomized smoothing, coincident with the empirical report that randomized smoothing leads to semantic gradients~\citep{kaur2019perceptually}.


Specifically, let $p^{\sigma}_{d}(x,y)$ be the data distribution augmented by a zero-mean Gaussian noise denoted as $\sqrt{\sigma}\omega$, where $\omega\sim\mathcal{N}(0,I)$ is a standard Gaussian and $p^{\sigma}_{d}(x,y)$ can be written as $p^{\sigma}_{d}(x,y)=\mathbb{E}_{\mathcal{N}(\omega;0,I)}\left[p_{d}(x-\sqrt{\sigma}\omega,y)\right]$.
We consider the Gaussian-augmented cross-entropy loss\footnote{We cannot write the loss in the form of KL divergence since there is $p^{\sigma}_{d}(x,y)\neq p^{\sigma}_{d}(x)p_{d}(y|x)$, as shown in Eq.~(\ref{eq26}).}
\begin{equation}
    \mathbf{R}_{G}(\theta;\sigma)=\mathbb{E}_{p^{\sigma}_{d}(x,y)}\left[-\log p_{\theta}(y|x)\right]\textrm{,}
\end{equation}
where $\mathbf{R}_{G}(\theta;0)$ degenerates to the cross-entropy loss. Then we derive the loss derivative of $\mathbf{R}_{G}(\theta;\sigma)$ w.r.t. $\sigma$ as
\begin{ftheorem}
\label{theorem5}
(Gradient Alignment) Given any model parameters $\theta$, the loss derivative of $\mathbf{R}_{G}(\theta;\sigma)$ w.r.t. $\sigma$ is
\begin{equation*}
\!\!\frac{d}{d\sigma}\mathbf{R}_{G}(\theta;\sigma)\!=\!\frac{1}{2}\mathbb{E}_{p^{\sigma}_{d}(x,y)}\!\left[\nabla_{x}\log p_{\theta}(y|x)^{\top}\nabla_{x}\log p^{\sigma}_{d}(x|y)\right]
\end{equation*}
where $p^{\sigma}_{d}(x|y)$ is defined as $\mathbb{E}_{\mathcal{N}(\omega;0,I)}\left[p_{d}(x-\sqrt{\sigma}\omega|y)\right]$.
\end{ftheorem}
In short, Theorem~\ref{theorem5} is proved following a similar routine as in \citet{lyu2009interpretation}. Particularly, for small values of $\sigma$, there is
\begin{equation}
    \mathbf{R}_{G}(\theta;\sigma)=\mathbf{R}_{G}(\theta;0)+\sigma\cdot\frac{d}{d\sigma}\mathbf{R}_{G}(\theta;\sigma)\Big|_{\sigma=0}+o(\sigma)\textrm{,}
\end{equation}
which says that Gaussian-augmented learning could be decomposed into standard learning (first term), and gradient alignment (second term) encouraging $\nabla_{x}\log p_{\theta}(y|x)$ to match the direction of $-\nabla_{x}\log p_{d}(x|y)$. This result adds to the evidence that robust learning has generative properties.

\begin{table}[t]
  \centering
  \footnotesize
  \setlength{\tabcolsep}{2.5pt}
  \vspace{-0.cm}
  \caption{Classification accuracy (\%) on clean images and under 10-steps PGD attack. Here we use a ResNet-18 model trained on CIFAR-10. We do not test under AutoAttack since this is only a qualitative study on effects of different losses used in PGD-AT. We find distance metrics do not work well in practice, while different losses adopt different suitable learning rates as highlighted.}
  \vspace{0.2cm}
  \renewcommand*{\arraystretch}{1.2}
    \begin{tabular}{ll|cc|cc|cc}
    \toprule
\multirow{2}{*}{Loss}  &  \multirow{2}{*}{Alias}   & \multicolumn{2}{c|}{$l.r.=0.1$} & \multicolumn{2}{c|}{$l.r.=0.05$} & \multicolumn{2}{c}{$l.r.=0.01$} \\
&   & Clean & PGD & Clean & PGD & Clean & PGD   \\
\midrule
$\|P-Q\|_{2}$ & $\ell_{2}$-dis.  & 75.91 & 52.16 & \multicolumn{1}{>{\columncolor{mycyan}}c}{77.98} & \multicolumn{1}{>{\columncolor{mycyan}}c}{52.74} & 78.45 & 51.13 \\
$\|P-Q\|_{1}$ & $\ell_{1}$-dis.  & 58.51 & 43.87 & 64.88 & 46.77 & \multicolumn{1}{>{\columncolor{mycyan}}c}{70.02} & \multicolumn{1}{>{\columncolor{mycyan}}c}{47.76} \\
$\|P-Q\|_{\infty}$ & $\ell_{\infty}$-dis.  & 58.34 & 43.71 & 59.75 & 45.02 & \multicolumn{1}{>{\columncolor{mycyan}}c}{65.65} & \multicolumn{1}{>{\columncolor{mycyan}}c}{46.36} \\
$\sqrt{\textrm{JS}(P\|Q)}$ & JS-dis.  & 53.06 & 40.08 & 55.27 & 41.86 & \multicolumn{1}{>{\columncolor{mycyan}}c}{68.50} & \multicolumn{1}{>{\columncolor{mycyan}}c}{46.49} \\
\midrule
$\textrm{JS}(P\|Q)$ & JS-div.  & 79.41 & 51.75 & \multicolumn{1}{>{\columncolor{mycyan}}c}{81.27} & \multicolumn{1}{>{\columncolor{mycyan}}c}{51.85} & 80.12 & 49.10 \\
$\textrm{KL}(P\|Q)$ & KL-div.  & \multicolumn{1}{>{\columncolor{mycyan}}c}{82.74} & \multicolumn{1}{>{\columncolor{mycyan}}c}{53.02} & 83.21 & 51.52 & 82.65 & 47.45\\
$\|P-Q\|_{1}^{2}$ & - & 79.87 & 50.96 & \multicolumn{1}{>{\columncolor{mycyan}}c}{81.49} & \multicolumn{1}{>{\columncolor{mycyan}}c}{52.00} & 81.26 & 47.51 \\
$\|P-Q\|_{2}^{2}$ & SE  & 80.59 & 54.63 & \multicolumn{1}{>{\columncolor{mycyan}}c}{83.38} & \multicolumn{1}{>{\columncolor{mycyan}}c}{54.01} & 81.43 & 51.13 \\
    \bottomrule
    \end{tabular}%
    \vspace{-0.0cm}
    \label{table3}
\end{table}%

\section{Experiments}
\label{sec6}
In this section we first discuss the 0-1 version of SCORE used for evaluation, to show the criterion is aligned with common practice; then, we demonstrate empirical results. Code is at \url{https://github.com/P2333/SCORE}.

\subsection{The 0-1 Version of SCORE for Evaluation}
\label{sec51}
In the test phase, we mostly apply 0-1 version of errors for evaluation~\citep{friedman2001elements}. Specifically, the 0-1 version of the standard error $\mathbf{R}_{\textrm{Standard}}(\theta)$ can be written as
\begin{equation}
    \mathbb{E}_{p_{d}(x)}\left[\mathbf{1}\left(\mathcal{Y}_{\theta}(x)\neq\mathcal{Y}_{d}(x)\right)\right]\textrm{.}
    \label{eq12}
\end{equation}
Recall that $\mathcal{Y}_{\theta}(x)$ and $\mathcal{Y}_{d}(x)$ are the hard labels obtained by taking $\argmax$ w.r.t. $p_{\theta}(y|x)$ and $p_{d}(y|x)$, respectively. Similarly, the 0-1 version of the robust error $\mathbf{R}_{\textrm{Madry}}(\theta)$ is
\begin{equation}
    \mathbb{E}_{p_{d}(x)}\left[\max_{{\color{red}x'}\in B(x)}\mathbf{1}\left(\mathcal{Y}_{\theta}({\color{red}x'})\neq\mathcal{Y}_{d}(x)\right)\right]\textrm{.}
    \label{eq13}
\end{equation}
Then one would naturally wonder: does SCORE induce a new evaluation criterion? Formally, the 0-1 version of $\mathbf{R}_{\textrm{SCORE}}(\theta)$ is formulated as
\begin{equation}
    \mathbb{E}_{p_{d}(x)}\left[\max_{{\color{red}x'}\in B(x)}\mathbf{1}\left(\mathcal{Y}_{\theta}({\color{red}x'})\neq\mathcal{Y}_{d}({\color{red}x'})\right)\right]\textrm{.}
    \label{eq14}
\end{equation}
Although SCORE advocates that $p_{d}(y|x')$ changes w.r.t. $x'$, the hard label $\mathcal{Y}_{d}(x')$ can be reasonably assumed to be invariant, i.e., $\mathcal{Y}_{d}(x')=\mathcal{Y}_{d}(x)$ for any $x'\in B(x)$. Actually this assumption is widely accepted in the adversarial community~\citep{biggio2013evasion,Goodfellow2014}. 
Given this, we conclude that the criteria in Eq.~(\ref{eq13}) and Eq.~(\ref{eq14}) are equivalent, and then $p_{\theta^*}(y|x)=p_{d}(y|x)$ is the optimal solution for both of them. Thus in our experiments, \emph{we do not need to modify the commonly used evaluation criterion}.

\subsection{Basic Setting without Extra or Generated Data}
In the basic setting, neither extra nor generated data are used.
We follow \citet{pang2020bag} and apply ResNet-18~\citep{he2016deep} as the model architecture. 
In training, we use SGD momentum optimizer with batch size $128$ and weight decay $5\times10^{-4}$.
We exploit the PGD-AT~\citep{madry2018towards} and TRADES~\citep{zhang2019theoretically} frameworks. 
The training attack used is 10-steps PGD with step size $\alpha=2/255$ for $\ell_{\infty}$ threat model and $\alpha=16/255$ for $\ell_{2}$ threat model.
The training runs for $110$ epochs with the learning rate decaying by a factor of $0.1$ at the $100$ and $105$ epoch, respectively. The hyperparameter $\beta=6$ in the TRADES experiments.

\textbf{Distance metric does not work well.} We first ablate on the effectiveness of different losses used in the PGD-AT objective $\mathbf{R}_{\textrm{Madry}}(\theta)$. 
As reported in Table~\ref{table3}, we choose $\ell_{1}$-, $\ell_{2}$-, $\ell_{\infty}$-, and JS distances, their squared variants, as well as KL divergence. 
We set three initial learning rates of $0.1$, $0.05$, and $0.01$. We evaluate under PGD attacks~\citep{madry2018towards} for a qualitative study. 
As can be seen, distance metrics do not work well in practice due to their sublinear property. 
Thus, we select squared error (SE) as a typical example developed by our analyses and compare effectiveness of SE-based instantiations with KL-based baselines.

\begin{table}[t]
  \centering
  \footnotesize
  \vspace{-0.cm}
  \caption{Classification accuracy (\%) on clean images and under AutoAttack ($\ell_{\infty}$, $\epsilon=8/255$). Here we use  ResNet-18 trained by PGD-AT or TRADES on CIFAR-10, using KL divergence or squared error (SE) as the loss function. Clipping loss is executed at every training step, compatible with early-stopping. We average the results over five runs and report the mean $\pm$ standard deviation.}
  \vspace{0.25cm}
  \renewcommand*{\arraystretch}{1.1}
    \begin{tabular}{ccccc}
    \toprule
Method & Loss & Clip & Clean & AutoAttack   \\
\midrule
 \multirow{3}{*}{PGD-AT} & KL div. & - & 82.46 $\pm$ 0.41 & 48.39 $\pm$ 0.14 \\
 & SE & \ding{55} & 82.13 $\pm$ 0.14 & 49.41 $\pm$ 0.27 \\
 & SE & \ding{51} & \textbf{82.80 $\pm$ 0.16} & \textbf{49.63 $\pm$ 0.17} \\
\midrule
  \multirow{3}{*}{TRADES} & KL div. & - & 81.47 $\pm$ 0.12 & 49.14 $\pm$ 0.16 \\
 & SE & \ding{55} & 83.50 $\pm$ 0.05 & 49.44 $\pm$ 0.35 \\
& SE & \ding{51} & \textbf{83.75 $\pm$ 0.14} & \textbf{49.57 $\pm$ 0.28} \\
    \bottomrule
    \end{tabular}%
    \vspace{-0.2cm}
    \label{table1}
\end{table}%

\textbf{SE outperforms KL divergence.} In Table~\ref{table1}, we substitute KL divergence in the objectives of PGD-AT and TRADES with SE. According to Table~\ref{table3}, we use $0.05$ initial learning rate for our methods and $0.1$ for baselines. For our methods, we report the results on the checkpoint with the highest value of PGD-10 (SE) accuracy on a separate validation set, similarly to \citet{rice2020overfitting}. The best checkpoint is selected for baselines by the highest value of PGD-10 (KL). Besides, we also introduce a clipping operation to the loss values. For PGD-AT, we choose the clipping threshold of $0.4$, and for TRADES, we choose $0.3$. We evaluate the model robustness under AutoAttack~\citep{croce2020reliable}. SE can improve clean accuracy and/or robustness for free, i.e., without extra computation.

\begin{table}[t]
  \centering
  \footnotesize
  \vspace{-0.cm}
  \caption{Classification accuracy (\%) on clean images and under AutoAttack ($\ell_{\infty}$, $\epsilon=8/255$). The model is WRN-28-10 (SiLU), following the training pipeline in \citet{rebuffi2021fixing} and using 1M DDPM generated data. KL divergence is substituted with the SE function in TRADES, and no clipping loss is executed.}
  \vspace{0.2cm}
  \renewcommand*{\arraystretch}{1.1}
    \begin{tabular}{cccc}
    \toprule
Dataset & $\beta$ & Clean  & AutoAttack \\
\midrule
\multirow{5}{*}{\textbf{CIFAR-10}} & 6 & 86.64 $\pm$ 0.13 & 60.78 $\pm$ 0.16\\
& 5 & 87.19 $\pm$ 0.20 & 61.05 $\pm$ 0.11\\
& 4 & 87.89 $\pm$ 0.19 & 61.11 $\pm$ 0.27\\
& 3 & 88.60 $\pm$ 0.13 & 60.89 $\pm$ 0.09\\
& 2 & 89.28 $\pm$ 0.15 & 60.13 $\pm$ 0.21\\
\midrule
\multirow{2}{*}{\textbf{CIFAR-100}} & 4 & 61.94 $\pm$ 0.13 & 31.21 $\pm$ 0.12\\
& 3 & 63.12 $\pm$ 0.37 & 31.01 $\pm$ 0.09\\
    \bottomrule
    
    \end{tabular}%
    \vspace{-0.4cm}
    \label{table2}
\end{table}%

\begin{table*}[t]
  \centering
  \footnotesize
  \vspace{-0.cm}
  \caption{Classification accuracy (\%) on clean images and under AutoAttack. The results of our methods are in \textbf{bold}, and no clipping loss is executed. Here $^\ddagger$ means \emph{no CutMix applied}, following~\citet{rade2021helperbased}. We use a batch size of 512 and train for 400 epochs due to limited resources, while a larger batch size of 1024 and training for 800 epochs are expected to achieve better performance.}
  \vspace{0.2cm}
  \renewcommand*{\arraystretch}{1.1}
    \begin{tabular}{clcccccc}
    \toprule
Dataset & Method & Architecture &  DDPM & Batch & Epoch & Clean & AutoAttack \\
\midrule
\multirow{12}{*}{{\begin{tabular}{c}\textbf{CIFAR-10}\\
    ($\ell_{\infty}$, $\epsilon=8/255$)
                \end{tabular}}} & \citet{rice2020overfitting} & WRN-34-20  & \ding{55} & 128 & 200 & 85.34 & 53.42 \\
                
& \citet{zhang2020attacks} & WRN-34-10  & \ding{55} & 128 & 120 & 84.52 & 53.51 \\

& \citet{pang2020bag} & WRN-34-20  & \ding{55} & 128 & 110 & 86.43 & 54.39 \\

& \citet{wu2020adversarial} & WRN-34-10  &\ding{55} & 128 & 200 & 85.36 & 56.17 \\

& \citet{gowal2020uncovering} & WRN-70-16  &\ding{55} & 512 & 200 & 85.29 & 57.14 \\

\cmidrule{2-8}

& \citet{rebuffi2021fixing}$^\ddagger$ & WRN-28-10  & 1M & 1024 & 800 & 85.97 & 60.73 \\

& \hspace{0.2cm} + \textbf{Ours} (KL $\rightarrow$ SE, $\beta=3$) & WRN-28-10  & 1M & 512 & 400 & \textbf{88.61} & \textbf{61.04} \\

& \hspace{0.2cm} + \textbf{Ours} (KL $\rightarrow$ SE, $\beta=4$) & WRN-28-10 & 1M & 512 & 400 & \textbf{88.10} & \textbf{61.51} \\

& \citet{rebuffi2021fixing}$^\ddagger$ & WRN-70-16 & 1M  & 1024 & 800 & 86.94 & 63.58 \\

& \hspace{0.2cm} + \textbf{Ours} (KL $\rightarrow$ SE, $\beta=3$) & WRN-70-16 & 1M &  512 & 400 & \textbf{89.01} & \textbf{63.35} \\

& \hspace{0.2cm} + \textbf{Ours} (KL $\rightarrow$ SE, $\beta=4$) & WRN-70-16 & 1M & 512 & 400 & \textbf{88.57} & \textbf{63.74} \\



\cmidrule{2-8}
& \citet{gowal2021improving} & WRN-70-16 & 100M & 1024 & 2000 & 88.74 & 66.10 \\

\midrule

\multirow{5}{*}{\begin{tabular}{c}\textbf{CIFAR-10}\\
    ($\ell_{2}$, $\epsilon=128/255$)
                \end{tabular}} & \citet{wu2020adversarial} & WRN-34-10 & \ding{55} & 128 & 200 & 88.51 & 73.66 \\
                
& \citet{gowal2020uncovering} & WRN-70-16 & \ding{55} & 512 & 200 & 90.90 & 74.50 \\

\cmidrule{2-8}

& \citet{rebuffi2021fixing}$^\ddagger$ & WRN-28-10 & 1M & 1024 & 800 & 90.24 & 77.37 \\

& \hspace{0.2cm} + \textbf{Ours} (KL $\rightarrow$ SE, $\beta=3$) & WRN-28-10 & 1M & 512 & 400 & \textbf{91.52} & \textbf{77.89} \\

& \hspace{0.2cm} + \textbf{Ours} (KL $\rightarrow$ SE, $\beta=4$) & WRN-28-10 & 1M & 512 & 400 & \textbf{90.83} & \textbf{78.10} \\

\midrule
\multirow{8}{*}{\begin{tabular}{c}\textbf{CIFAR-100}\\
    ($\ell_{\infty}$, $\epsilon=8/255$)
                \end{tabular}} & \citet{wu2020adversarial} & WRN-34-10 & \ding{55} & 128 & 200 & 60.38 & 28.86 \\
                
& \citet{gowal2020uncovering} & WRN-70-16 & \ding{55} & 512 & 200 & 60.86 & 30.03 \\

\cmidrule{2-8}

& \citet{rebuffi2021fixing}$^\ddagger$ & WRN-28-10 & 1M & 1024 & 800 & 59.18 & 30.81 \\

& \hspace{0.2cm} + \textbf{Ours} (KL $\rightarrow$ SE, $\beta=3$) & WRN-28-10 & 1M & 512 & 400 & \textbf{63.66} & \textbf{31.08} \\

& \hspace{0.2cm} + \textbf{Ours} (KL $\rightarrow$ SE, $\beta=4$) & WRN-28-10 & 1M &  512 & 400 & \textbf{62.08} & \textbf{31.40} \\

& \citet{rebuffi2021fixing}$^\ddagger$ & WRN-70-16 & 1M & 1024 & 800 & 60.46 & 33.49 \\

& \hspace{0.2cm} + \textbf{Ours} (KL $\rightarrow$ SE, $\beta=3$) & WRN-70-16 & 1M & 512 & 400 & \textbf{65.56} & \textbf{33.05} \\

& \hspace{0.2cm} + \textbf{Ours} (KL $\rightarrow$ SE, $\beta=4$) & WRN-70-16 & 1M & 512 & 400 & \textbf{63.99} & \textbf{33.65} \\

    \bottomrule

    \end{tabular}%
    \vspace{-0.25cm}
    \label{table4}
\end{table*}%

\subsection{Advanced Setting with DDPM Generated Data}
Recent progress shows that generative models trained solely on the original training set can be leveraged to improve model robustness~\citep{rebuffi2021fixing,gowal2021improving,sehwag2021improving}. 
We thus follow the setting of \citet{rebuffi2021fixing} and its PyTorch implementation\footnote{https://github.com/imrahulr/adversarial\_robustness\_pytorch}, using the provided 1M DDPM~\citep{ho2020denoising} generated data. For the sake of completeness, we briefly recap the training setup here. We use SiLU activation function~\citep{hendrycks2016gaussian} with WideResNet (WRN) backbones~\citep{zagoruyko2016wide}. We adopt weight averaging~\citep{izmailov2018averaging} with $\tau=0.995$. 
For optimizer we use SGD with Nesterov momentum~\citep{nesterov1983method}, momentum factor being $0.9$ and weight decay $5\times10^{-4}$.
We further use cyclic learning rates~\citep{smith2019super} with cosine annealing. 
We use a batch size of 512 and train for 400 epochs (with an initial learning rate of $0.2$) due to limited computational resources; a larger batch size of 1024 and longer training of 800 epochs (with an initial learning rate of $0.4$) could further improve our method, as employed in~\citet{rebuffi2021fixing}.
The original-to-generated data ratio in each batch is $0.3$ on CIFAR-10 and $0.4$ on CIFAR-100~\citep{gowal2021improving}.

\textbf{Values of $\beta$.} In Table~\ref{table2}, we study the effect of $\beta$ in TRADES on the robustness-accuracy trade-off. 
Unlike previous observations~\citep{zhang2019theoretically}, we find when using large models with extra generated data, smaller values of $\beta$ (i.e., $3$ or $4$) facilitate higher clean accuracy while keeping robust accuracy under AutoAttack almost unchanged.
Other more advanced ways for tuning $\beta$ can also be applied~\citep{wu2021do}. Theoretically, Theorem~\ref{theorem2} tells us that smaller values of $\beta$ make TRADES more aligned with PGD-AT.

\textbf{Comparison with state-of-the-art.} In Table~\ref{table4}, we consider threat models of ($\ell_{\infty}$, $\epsilon=8/255$) and ($\ell_{2}$, $\epsilon=128/255$) on CIFAR-10, and ($\ell_{\infty}$, $\epsilon=8/255$) on CIFAR-100. We resort to RobustBench~\citep{croce2020robustbench} for comparing with the top-performance robust models. We run on the model architectures of WRN-28-10 and WRN-70-16. Note that we do not apply \emph{CutMix} following \citep{rade2021helperbased}, since we find the effectiveness of CutMix may rely on the specific implementation of parallel computing and requires further exploration. As observed, simply substituting KL divergence with SE and using a relatively small value of $\beta$ can significantly improve clean accuracy, with comparable or even better robustness.

\section{Conclusion and Discussion}
We attribute the trade-off between robustness and accuracy to the improper definition of robustness. Essentially, the robust error $\mathbf{R}_{\textrm{Madry}}(\theta)$ is a surrogate objective of its 0-1 form, but the problem is that it unintentionally converts the hard-label invariance (i.e., $\mathcal{Y}_{d}(x')$ is unchanged) into the distributional invariance (i.e., $p_{d}(y|x')$ should equal $p_{d}(y|x)$) for $\forall x'\in B(x)$. The former is a reasonable inductive bias, while the latter is an overcorrection towards smoothness. Thus we propose SCORE and provide an efficient way to optimize it. SCORE brings us many new insights for explaining the phenomena of overfitting and semantic gradients encountered on robust models. Inspired by SCORE, substituting KL divergence with SE effectively alleviates the empirical robustness-accuracy trade-off.

\textbf{Empirical trade-off still exists.} SCORE ensures that robustness and accuracy are reconcilable in the sense of taking expectation w.r.t. $p_{d}$. However, in the finite-sample cases, the insufficiency of training data could cause empirical trade-off~\citep{schmidt2018adversarially}. This effect is also observed in Fig.~\ref{fig:2} when using the standard error or SCORE. Fortunately, SCORE promises that the trained model will finally converge to a self-consistent solution as more data are collected.

\textbf{$B(x)$ can be arbitrary in SCORE.} The local invariance assumed in $\mathbf{R}_{\textrm{Madry}}(\theta)$ constrains the allowed set $B(x)$ to be local around $x$ (e.g., $\ell_{p}$-balls). In contrast, SCORE exploits equivariance instead of invariance; thus, $B(x)$ could be an arbitrary set (e.g., involving the points with similar semantics as $x$), whereas most of the conclusions proved for SCORE still hold. Given this, we can apply SCORE to a wider range of tasks, with guarantees of self-consistency and enjoying sample efficiency brought by robust optimization.

\section*{Acknowledgements}
{The authors would like to thank Kun Xu for the insightful discussion on the idea of SCORE proposed in this paper. This work was supported by National Key Research and Development Program of China (Nos. 2020AAA0104304, 2020AAA0106302, 2017YFA0700904), NSFC Projects (Nos. 62061136001, 61621136008, 62076147, U19B2034, U19A2081, U1811461),  Tsinghua-Huawei Joint Research Program, a grant from Tsinghua Institute for Guo Qiang, and the High Performance Computing Center, Tsinghua University. Tianyu Pang was partly supported by Microsoft Research Asia Fellowship and Baidu Scholarship.}

\bibliographystyle{plainnat}
\bibliography{main}

\clearpage
\appendix
\onecolumn
\section{Proofs}
\label{proofs}
In this section, we provide proofs of the Theorems proposed in the main text.

\subsection{Proof of Theorem~\ref{theorem1}}
\label{proof1}
According to the symmetry and the triangle inequality of any distance metric $\mathcal{D}(\cdot||\cdot)$, we have
\begin{equation}
\begin{split}
         &\mathbf{R}_{\textrm{SCORE}}^{\mathcal{D}}(\theta)\\
         =&\mathbb{E}_{p_{d}(x)}\left[\max_{x'\in B(x)}\mathcal{D}\left(p_{d}(y|x')\big\|p_{\theta}(y|x')\right)\right]\\
         \leq&\mathbb{E}_{p_{d}(x)}\left[\max_{x'\in B(x)}\left(\mathcal{D}\left(p_{d}(y|x')\big\|p_{d}(y|x)\right)+\mathcal{D}\left(p_{d}(y|x)\big\|p_{\theta}(y|x')\right)\right)\right]\\
         \leq&\mathbb{E}_{p_{d}(x)}\left[\max_{x'\in B(x)}\mathcal{D}\left(p_{d}(y|x')\big\|p_{d}(y|x)\right)+\max_{x'\in B(x)}\mathcal{D}\left(p_{d}(y|x)\big\|p_{\theta}(y|x')\right)\right]\\
         =&\mathbb{E}_{p_{d}(x)}\left[\max_{x'\in B(x)}\mathcal{D}\left(p_{d}(y|x)\big\|p_{\theta}(y|x')\right)\right] + C^{\mathcal{D}} \\
          =&\mathbf{R}_{\textrm{Madry}}^{\mathcal{D}}(\theta) + C^{\mathcal{D}}\textrm{.}
\end{split}
\end{equation}
Furthermore, we can show that
\begin{equation}
    \begin{split}
         &\mathbf{R}_{\textrm{SCORE}}^{\mathcal{D}}(\theta) + C^{\mathcal{D}}\\
         =&\mathbb{E}_{p_{d}(x)}\left[\max_{x'\in B(x)}\mathcal{D}\left(p_{d}(y|x')\big\|p_{\theta}(y|x')\right)+\max_{x'\in B(x)}\mathcal{D}\left(p_{d}(y|x')\big\|p_{d}(y|x)\right)\right]\\
         \geq&\mathbb{E}_{p_{d}(x)}\left[\max_{x'\in B(x)}\left(\mathcal{D}\left(p_{d}(y|x')\big\|p_{d}(y|x)\right)+\mathcal{D}\left(p_{d}(y|x')\big\|p_{\theta}(y|x')\right)\right)\right]\\
         \geq&\mathbb{E}_{p_{d}(x)}\left[\max_{x'\in B(x)}\mathcal{D}\left(p_{d}(y|x)\big\|p_{\theta}(y|x')\right)\right] \\
          =&\mathbf{R}_{\textrm{Madry}}^{\mathcal{D}}(\theta)\textrm{.}
\end{split}
\end{equation}
Similarly, there is $\mathbf{R}_{\textrm{Madry}}^{\mathcal{D}}(\theta) + \mathbf{R}_{\textrm{SCORE}}^{\mathcal{D}}(\theta) \geq C^{\mathcal{D}}$, thus in conclusion, we prove that
\begin{equation}
    \lvert{\mathbf{R}_{\textrm{Madry}}^{\mathcal{D}}(\theta) - C^{\mathcal{D}}}\rvert\leq \mathbf{R}_{\textrm{SCORE}}^{\mathcal{D}}(\theta) \leq \mathbf{R}_{\textrm{Madry}}^{\mathcal{D}}(\theta) + C^{\mathcal{D}}\textrm{.}
\end{equation}
\qed

\subsection{Proof of Theorem~\ref{theorem3}}
According to Theorem~\ref{theorem1}, it is easy to show that
\begin{equation}
\label{eq19}
    \lvert{\mathbf{R}_{\textup{SCORE}}^{\mathcal{D}}(\theta)-C^{\mathcal{D}}\rvert} \leq \mathbf{R}_{\textup{Madry}}^{\mathcal{D}}(\theta)\textrm{,}
\end{equation}
Since $\phi(\cdot)$ is monotonically increasing, there is
\begin{equation}
    \argmax_{{x'}\in B(x)}\mathcal{D}\left(p_{d}(y|x)\big\|p_{\theta}(y|{x'})\right)=\argmax_{{x'}\in B(x)}\phi\circ\mathcal{D}\left(p_{d}(y|x)\big\|p_{\theta}(y|{x'})\right)\textrm{.}
\end{equation}
Additionally $\phi(\cdot)$ is convex, then by Jensen's inequality, we have
\begin{equation}
    \phi\left(\mathbf{R}_{\textup{Madry}}^{\mathcal{D}}(\theta)\right)\leq \mathbf{R}_{\textup{Madry}}^{\phi\circ\mathcal{D}}(\theta)\textrm{.}
\end{equation}
Take this formula into Eq.~(\ref{eq19}), we prove that
\begin{equation}
    \lvert{\mathbf{R}_{\textup{SCORE}}^{\mathcal{D}}(\theta)-C^{\mathcal{D}}\rvert} \leq \phi^{-1}\left(\mathbf{R}_{\textup{Madry}}^{\phi\circ\mathcal{D}}(\theta)\right)\textrm{.}
\end{equation}
\qed

\subsection{Proof of Theorem~\ref{theorem2}}
As to the TRADES objective, we have
\begin{equation}
\begin{split}
         &\mathbf{R}_{\textrm{TRADES}}^{\mathcal{D}}(\theta;\beta)\\
         =&\mathbb{E}_{p_{d}(x)}\left[\mathcal{D}\left(p_{d}(y|x)\big\|p_{\theta}(y|x)\right)+\beta\cdot\max_{x'\in B(x)}\mathcal{D}\left(p_{\theta}(y|x)\big\|p_{\theta}(y|x')\right)\right]\\
         \leq&\mathbb{E}_{p_{d}(x)}\left[\mathcal{D}\left(p_{d}(y|x)\big\|p_{\theta}(y|x)\right)+\beta\cdot\max_{x'\in B(x)}\left(\mathcal{D}\left(p_{d}(y|x)\big\|p_{\theta}(y|x)\right)+\mathcal{D}\left(p_{d}(y|x)\big\|p_{\theta}(y|x')\right)\right)\right]\\
         =&\mathbb{E}_{p_{d}(x)}\left[(1+\beta)\cdot\mathcal{D}\left(p_{d}(y|x)\big\|p_{\theta}(y|x)\right)+\beta\cdot\max_{x'\in B(x)}\mathcal{D}\left(p_{d}(y|x)\big\|p_{\theta}(y|x')\right)\right]\\
         \leq&\mathbb{E}_{p_{d}(x)}\left[(1+2\beta)\cdot\max_{x'\in B(x)}\mathcal{D}\left(p_{d}(y|x)\big\|p_{\theta}(y|x')\right)\right]\\
         =&(1+2\beta)\cdot\mathbf{R}_{\textrm{Madry}}^{\mathcal{D}}(\theta)
\end{split}
\label{eq18}
\end{equation}
Besides, it is easy to show that $\mathbf{R}_{\textrm{TRADES}}^{\mathcal{D}}(\theta;\beta)\geq\mathbf{R}_{\textrm{Madry}}^{\mathcal{D}}(\theta)$ holds for any $\beta\geq 1$.
\qed

\subsection{Proof of Theorem~\ref{theorem4}}
In the $\ell_{p}$-norm threat model of $B(x)=\{x'\big|\|x'-x\|_{p}\leq \epsilon\}$, we take the first-order expansion similar as in \citet{simon2019first}, there is
\begin{equation}
        \mathbf{R}_{\textrm{SCORE}}^{\mathcal{D}}(\theta) - \mathbf{R}_{\textrm{Standard}}^{\mathcal{D}}(\theta)=\epsilon\cdot\mathbb{E}_{p_{d}(x)}\left[\Big\|\nabla_{x}\mathcal{D}\left(p_{d}(y|x)\big\|p_{\theta}(y|x)\right)\Big\|_{q}\right] + o(\epsilon)\textrm{.}
\end{equation}
In particular, when $\mathcal{D}$ is the $\ell_{1}$-distance metric, we have
\begin{equation}
        \mathbf{R}_{\textrm{SCORE}}^{\ell_{1}}(\theta) - \mathbf{R}_{\textrm{Standard}}^{\ell_{1}}(\theta)=\epsilon\cdot\mathbb{E}_{p_{d}(x)}\left[\Big\|\sum_{y}s_{y}\cdot\left(\nabla_{x}p_{d}(y|x)-\nabla_{x}p_{\theta}(y|x)\right)\Big\|_{q}\right] + o(\epsilon)\textrm{,}
    \label{eq17}
\end{equation}
where $s_{y}=\sgn(p_{d}(y|x)-p_{\theta}(y|x))$. Under the assumption, there is $s_{y}=1$ for $y=\mathcal{Y}_{d}(x)$; otherwise $s_{y}=-1$. Besides, since $\sum_{y}p_{d}(y|x)=\sum_{y}p_{\theta}(y|x)=1$, we know that
\begin{equation}
    \sum_{y}\nabla_{x}p_{d}(y|x)=\sum_{y}\nabla_{x}p_{\theta}(y|x)=0\textrm{.}
\end{equation}
Take these into Eq.~(\ref{eq17}), we can derive that
\begin{equation}
    \mathbf{R}_{\textrm{SCORE}}^{\ell_{1}}(\theta) - \mathbf{R}_{\textrm{Standard}}^{\ell_{1}}(\theta)=2\epsilon\cdot\mathbb{E}_{p_{d}(x)}\left[\left\|\nabla_{x}p_{d}(\mathcal{Y}_{d}(x)|x)-\nabla_{x}p_{\theta}(\mathcal{Y}_{d}(x)|x)\right\|_{q}\right] + o(\epsilon)\textrm{.}
\end{equation}
\qed

\subsection{Proof of Theorem~\ref{theorem5}}
Under Gaussian augmentation, the augmented joint distribution of $(x,y)$ becomes
\begin{equation}
\begin{split}
    p^{\sigma}_{d}(x,y)&=\int \mathcal{N}(\omega;0,I)\cdot p_{d}(x-\sqrt{\sigma}\omega,y)d\omega\\
    &=\mathbb{E}_{\mathcal{N}(\omega;0,I)}\left[p_{d}(x-\sqrt{\sigma}\omega|y)\right]p_{d}(y)\\
    &=p^{\sigma}_{d}(x|y)p_{d}(y)\\
    &\neq p^{\sigma}_{d}(x)p_{d}(y|x)\text{.}
\end{split}
\label{eq26}
\end{equation}

Then we can derive the derivatives of augmented cross-entropy loss w.r.t. $\sigma$ as
\begin{equation}
    \begin{split}
        &\frac{d}{d\sigma}\mathbb{E}_{p_{d}(y)}\mathbb{E}_{p^{\sigma}_{d}(x|y)}\left[-\log p_{\theta}(y|x)\right]\\
        =&-\mathbb{E}_{p_{d}(y)}\int \frac{d}{d\sigma}p^{\sigma}_{d}(x|y)\log p_{\theta}(y|x)dx \\
        =&-\frac{1}{2}\mathbb{E}_{p_{d}(y)}\int  \nabla_{x}^{\top}\nabla_{x} p^{\sigma}_{d}(x|y)\log p_{\theta}(y|x)dx \hspace{6cm}\textrm{(diffusion equation)}\\
        =&-\frac{1}{2}\mathbb{E}_{p_{d}(y)}\left[\sum_{i} \frac{\partial}{\partial x_{i}}p^{\sigma}_{d}(x|y)\log p_{\theta}(y|x)\Big|_{-\infty}^{\infty}-\int \nabla_{x}\log p_{\theta}(y|x)^{\top}\nabla_{x} p^{\sigma}_{d}(x|y)dx\right]\hspace{0.5cm}\textrm{(integration by parts)}\\
         =&\frac{1}{2}\mathbb{E}_{p_{d}(y)}\mathbb{E}_{p^{\sigma}_{d}(x|y)}\left[\nabla_{x}\log p_{\theta}(y|x)^{\top}\nabla_{x}\log p^{\sigma}_{d}(x|y)\right]\text{.}
    \end{split}
\end{equation}
The term $\sum_{i} \frac{\partial}{\partial x_{i}}p^{\sigma}_{d}(x|y)\log p_{\theta}(y|x)\Big|_{-\infty}^{\infty}$ equals to zero under the boundary assumption~\citep{hyvarinen2005estimation}. 
\qed

\subsection{Proof of Corollary~\ref{corollary1}}
Let $\phi(\cdot)=(\cdot)^{2}$ be the square function, which is monotonically increasing and convex. Take $\phi(\cdot)^{-1}=\sqrt{(\cdot)}$ and $\mathcal{D}$ be $\ell_{1}$-distance into Theorem~\ref{theorem3}, we have
\begin{equation}
    \lvert{\mathbf{R}_{\textup{SCORE}}^{\ell_{1}}(\theta)-C^{\ell_{1}}\rvert} \leq \sqrt{\mathbf{R}_{\textup{Madry}}^{\ell_{1}^{2}}(\theta)}\textrm{.}
\end{equation}
From the Pinsker's inequality we have
\begin{equation}
    \max_{x'\in B(x)}\|p_{d}(y|x)-p_{\theta}(y|x')\|_{1}^{2}\leq 2\cdot\max_{x'\in B(x)}{\mathrm{KL}(p_{d}(y|x)||p_{\theta}(y|x'))}\textrm{.}
\end{equation}
After taking expectation w.r.t. $p_{d}(x)$, we achieve that $\mathbf{R}_{\textup{Madry}}^{\ell_{1}^{2}}(\theta)\leq 2\cdot\mathbf{R}_{\textup{Madry}}(\theta)$, which finally proves that
\begin{equation}
    \lvert{\mathbf{R}_{\textup{SCORE}}^{\ell_{1}}(\theta)-C^{\ell_{1}}\rvert} \leq \sqrt{2\cdot\mathbf{R}_{\textup{Madry}}(\theta)}\textrm{.}
\end{equation}
\qed

\section{Detailed Derivations}
In this section, we provide detailed derivations to better support our conclusions.

\subsection{The Connection between \texorpdfstring{$\textbf{R}_{\textrm{Madry}}(\theta)$}{TEXT} and the Original Objective in \citet{madry2018towards}}
\label{Rmardy}
In \citet{madry2018towards}, the robust error $\mathbf{R}_{\textrm{Madry}}(\theta)$ is defined w.r.t. the a loss function $\mathcal{L}(x,y;\theta)$, e.g., the cross-entropy loss, formulated as:
\begin{equation}
     \widetilde{\mathbf{R}}_{\textrm{Madry}}(\theta)=\mathbb{E}_{p_{d}(x,y)}\left[\max_{x'\in B(x)}\mathcal{L}(x',y;\theta)\right]=\mathbb{E}_{p_{d}(x)}\left[\sum_{y}p_{d}(y|x)\max_{x'\in B(x)}\mathcal{L}(x',y;\theta)\right]\textrm{,}
\label{eq1}
\end{equation}
where the maximization w.r.t. $x'$ takes place \emph{before} the summation w.r.t. $y$. By setting $\mathcal{L}(x,y;\theta)=-\log p_{\theta}(y|x)$ be the cross-entropy loss, we can rewrite our defined $\mathbf{R}_{\textup{Madry}}$ as
\begin{equation}
\begin{split}
    \mathbf{R}_{\textup{Madry}}(\theta)=&\mathbb{E}_{p_{d}(x)}\left[\max_{x'\in B(x)}\textrm{KL}\left(p_{d}(y|x)\big\|p_{\theta}(y|x')\right)\right]\\
    =&\mathbb{E}_{p_{d}(x)}\left[\max_{x'\in B(x)}\sum_{y}p_{d}(y|x)\mathcal{L}(x',y;\theta)\right]+\underbrace{\mathbb{E}_{p_{d}(x)}\left[\sum_{y}p_{d}(y|x)\log p_{d}(y|x)\right]}_{\textbf{independent of } \theta}\textrm{,}
\end{split}
\end{equation}
where only the first term works when optimized w.r.t. $\theta$. As seen, in this formula the maximization w.r.t. $x'$ takes place \emph{after} the summation w.r.t. $y$. Although the summation operator $\sum_{y}$ and the maximization operator $\max_{x'\in B(x)}$ are not directly permutable, we can permute summation and gradient operators when we use first-order optimization methods, e.g, SGD or Adam. So the (first-order) updating directions of minimizing $\widetilde{\mathbf{R}}_{\textup{Madry}}(\theta)$ and $\mathbf{R}_{\textup{Madry}}(\theta)$ can be regarded as equal, and we treat them as the same objective in the main text.

\subsection{Using Score-based Learning to Optimize \texorpdfstring{$\textbf{R}_{\textrm{SCORE}}(\theta)$}{TEXT}}
\label{secB1}
To directly minimize SCORE with first-order optimizers, we need to explicitly compute $\nabla_{x}\textrm{KL}\left(p_{d}(y|x)\big\|p_{\theta}(y|x)\right)$ as
\begin{equation}
    \begin{split}
        &\nabla_{x}\textrm{KL}\left(p_{d}(y|x)\big\|p_{\theta}(y|x)\right)\\
        =&\sum_{y\in[L]}\left[-p_{d}(y|x)\nabla_{x}\log p_{\theta}(y|x)+\nabla_{x}p_{d}(y|x)\left(1+\log p_{d}(y|x)-\log p_{\theta}(y|x)\right)\right]\\
        =&\sum_{y\in[L]}\left[-p_{d}(y|x)\nabla_{x}\log p_{\theta}(y|x)+\nabla_{x}p_{d}(y|x)\left(\log p_{d}(y|x)-\log p_{\theta}(y|x)\right)\right]\\
        =&\sum_{y\in[L]}p_{d}(y|x)\big[-\underbrace{\nabla_{x}\log p_{\theta}(y|x)}_\textbf{model gradient}+\underbrace{\nabla_{x}\log p_{d}(y|x)}_\textbf{data gradient}\left(\log p_{d}(y|x)-\log p_{\theta}(y|x)\right)\big]\textrm{.}
    \end{split}
\end{equation}
As seen, we need to have access to the data gradient $\nabla_{x}\log p_{d}(y|x)$. To this end, score matching methods may be applied~\citep{vincent2011connection,song2019sliced,pang2020efficient}, based on the decomposition that
\begin{equation}
    \nabla_{x}\log p_{d}(y|x) = \nabla_{x}\log p_{d}(x|y) - \nabla_{x}\log p_{d}(x)\textrm{,}
\end{equation}
which means that we can estimate a conditional data score $\nabla_{x}\log p_{d}(x|y)$ and an unconditional data score $\nabla_{x}\log p_{d}(x)$ to obtain $\nabla_{x}\log p_{d}(y|x)$. We can use either an energy-based model (EBM) with back-propagation or a direct scorenet~\citep{song2019generative} as the data-score model. In our initial experiments, we employ NCSN++~\citep{song2019improving} to estimate the data scores, using denoising score matching (DSM)~\citep{vincent2011connection}. Even though MCMC methods like SGLD can generate high-quality images with the learned scores, we find that the learned scores are of high variance when applied into the discriminative learning process. Therefore, we do not further explore this pipeline in this paper, but we still regard score-based learning as a promising and principled way to optimize SCORE (just as the abbreviation implies).

\section{Detailed Discussion on the Effects of Randomized Smoothing}
\label{secC}
Recall that the Gaussian-augmented cross-entropy loss is defined as
\begin{equation}
    \mathbf{R}_{G}(\theta;\sigma)=\mathbb{E}_{ p^{\sigma}_{d}(x,y)}\left[-\log p_{\theta}(y|x)\right]\textrm{,}
\end{equation}
where $\mathbf{R}_{G}(\theta;0)$ is the cross-entropy loss. Then beyond the conclusion in Theorem~\ref{theorem5}, we can tune the effect of the gradient alignment term in $\mathbf{R}_{G}(\theta;\sigma)$ by subtracting a scaled $\mathbf{R}_{G}(\theta;0)$, controlled via an extra hyperparameter $\gamma$ as
\begin{equation}
    \begin{split}
        &\frac{1}{1-\gamma}\cdot\left[\mathbf{R}_{G}(\theta;\sigma)-\gamma\cdot\mathbf{R}_{G}(\theta;0)\right]\\
        =&\mathbf{R}_{G}(\theta;0)+\frac{\sigma}{2(1-\gamma)}\cdot\mathbb{E}_{ p^{\sigma}_{d}(x,y)}\left[\nabla_{x}\log p_{\theta}(y|x)^{\top}\nabla_{x}\log p^{\sigma}_{d}(x|y)\right]+o(\sigma)\text{.}
    \end{split}
    \label{appeq31}
\end{equation}
Note that the coefficient of the gradient alignment term is now $\frac{\sigma}{2(1-\gamma)}$. Thus, we can magnify the coefficient by increasing $\gamma$ close to one, rather than increasing $\sigma$ which could degrade clean performance. Besides, we can adaptively anneal $\gamma$ and $\sigma$ during training to keep $\frac{\sigma}{2(1-\gamma)}$ as a constant, while the saliency map of the trained models is more interpretable. In the implementation, $\sigma$ can be uniformly sampled for each data point, and $\gamma$ could point-wisely adapt to the value of $\sigma$. Similarly, we can also construct
\begin{equation}
    \begin{split}
    &\frac{1}{\gamma-1}\cdot\left[\gamma\cdot\mathbf{R}_{G}(\theta;0)-\mathbf{R}_{G}(\theta;\sigma)\right]\\
        =&\mathbf{R}_{G}(\theta;0)-\frac{\sigma}{2(\gamma-1)}\cdot\mathbb{E}_{ p^{\sigma}_{d}(x,y)}\left[\nabla_{x}\log p_{\theta}(y|x)^{\top}\nabla_{x}\log p^{\sigma}_{d}(x|y)\right]+o(\sigma)\text{.}
    \end{split}
    \label{appeq32}
\end{equation}
Interestingly, by comparing the gradient alignment terms in Eq.~(\ref{appeq31}) and Eq.~(\ref{appeq32}), the sign is reversed, i.e., we can even control the direction of gradient alignment.

\textbf{The gradient alignment cannot be perfectly achieved.} Since $\sum_{y}p_{\theta}(y|x)=1$, there is $\sum_{y}\nabla_{x}p_{\theta}(y|x)=0$. On the other hand, there is $\sum_{y}p_{d}(x|y)p_{d}(y)=p_{d}(x)$ and thus $\sum_{y}p_{d}(y)\nabla_{x}p_{d}(x|y)=\nabla_{x}p_{d}(x)$, which indicates that $p_{\theta}(y|x)$ cannot perfectly align with $p_{d}(x|y)$ or $-p_{d}(x|y)$ especially when $p_{d}(y)$ is an uniform distribution.

\textbf{Why does the gradient alignment come from training rather than inference?} Except for the Gaussian-augmented training, the most critical characteristic of randomized smoothing is to apply an ensemble of Gaussian-perturbed predictions during inference, formulated as
\begin{equation}
    p_{\theta}^{\sigma}(y|x)=\mathbb{E}_{\mathcal{N}(\omega;0,I)}\left[p_{\theta}(y|x+\sqrt{\sigma}\omega)\right]\textrm{.}
\end{equation}
In practice, the expectation is approximated by $N$ samples of $\omega_{1},\cdots,\omega_{N}$. So it is probable that the inference mechanism causes the semantic gradients. However, we notice that in Fig. 14 of \citet{kaur2019perceptually}, the authors ablate on the number of Monte Carlo samples used for gradient estimation. As observed, even for $N=1$ (i.e., using a single Gaussian-perturbed prediction), the gradients look perceptually aligned, which is certainly not the case for a standardly trained model (i.e., without Gaussian augmentation). Thus we attribute the semantic phenomenon more to the training phase.

\section{Additional Experiments}
In this section, we provide more technical details and empirical results.

\begin{figure}[t]
\begin{center}
\vspace{-0.cm}
\includegraphics[width=1.00\columnwidth]{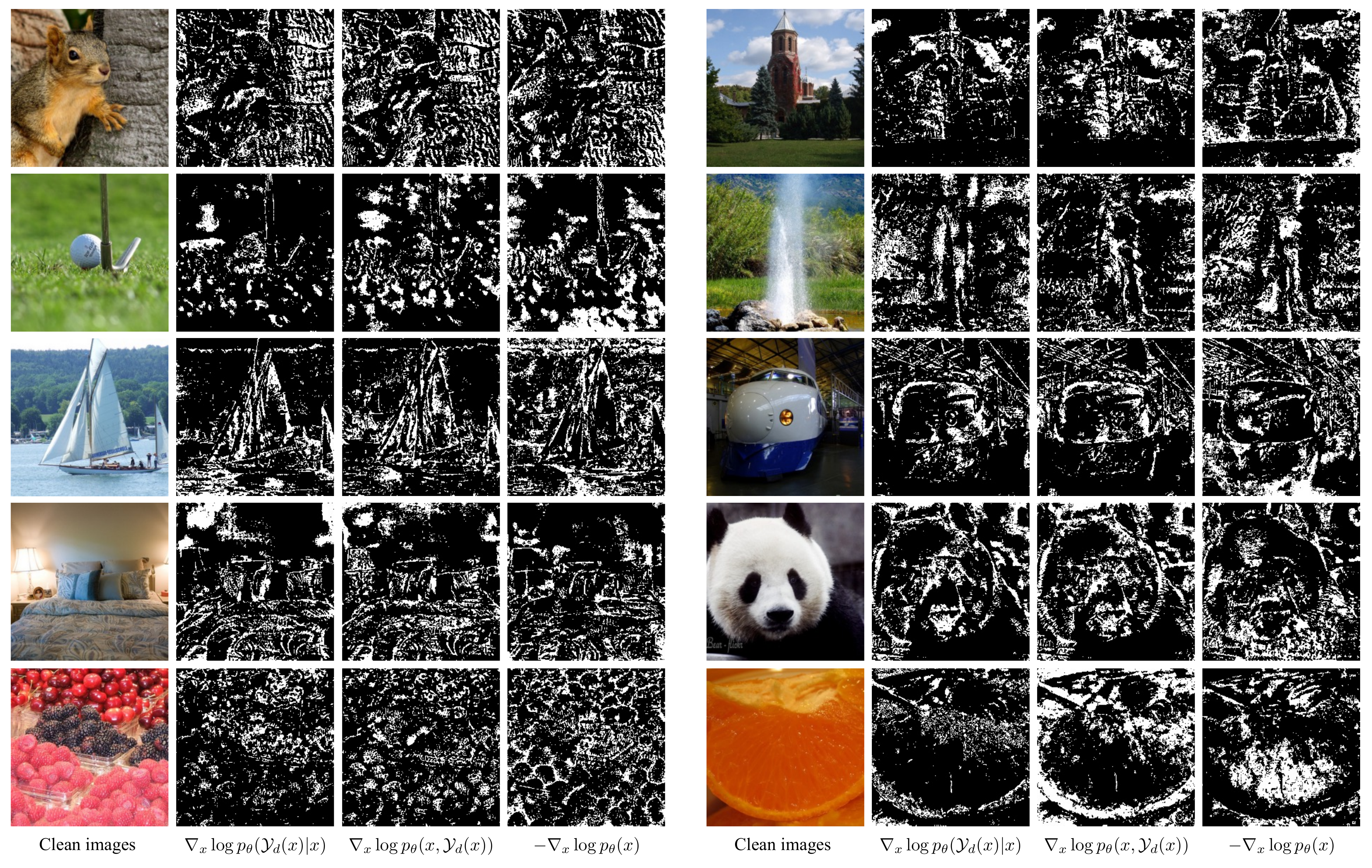}
\vspace{-0.7cm}
\caption{More examples of visualizing the semantic gradients of a ResNet-50 trained by FreeAT on ImageNet, using the same pipeline as in Fig.~\ref{fig:3}. The gradients will be noisy and almost uniformly scattered for standardly trained models (not plotted here).}
\label{appendixfig:1}
\end{center}
\vspace{0.3cm}
\end{figure}

\subsection{Visualization of (KL-based) Overfitting}
\label{secD2}
\begin{figure}[t]
\begin{center}
\vspace{-0.cm}
\includegraphics[width=1.00\columnwidth]{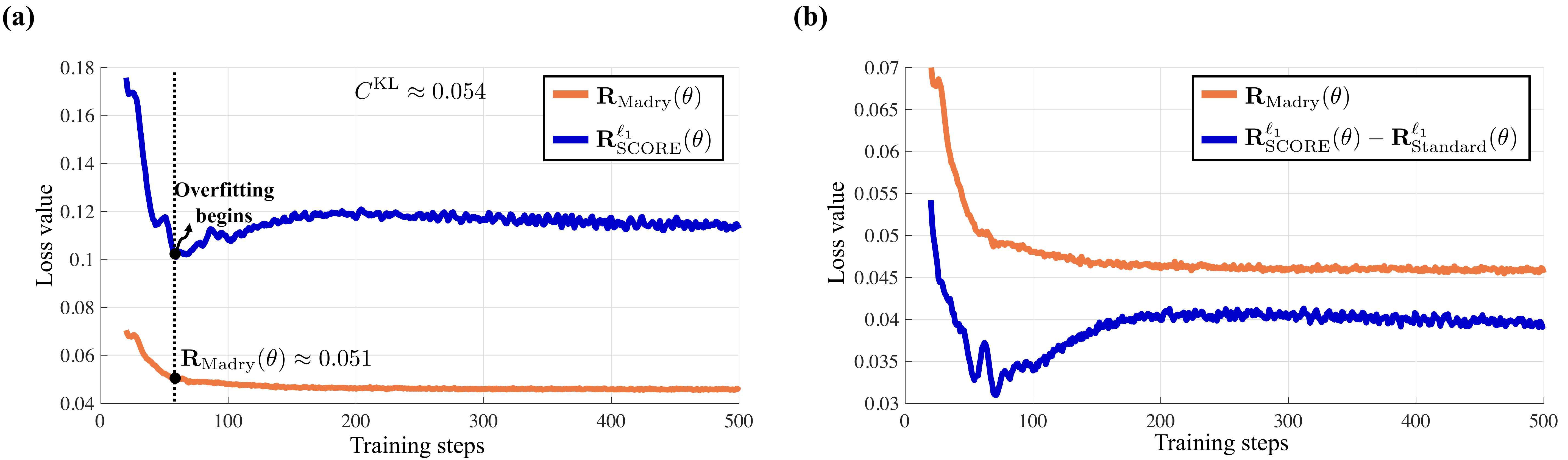}
\vspace{-0.5cm}
\caption{Toy demo with the same settings as in Fig.~\ref{fig:1}. \textbf{(a)} Illustration of Corollary~\ref{corollary1} on the overfitting phenomenon. $\mathbf{R}_{\textup{SCORE}}^{\ell_{1}}(\theta)$ begins to overfit when $\mathbf{R}_{\textup{Madry}}(\theta)$ is minimized to around $C^{\textrm{KL}}$; \textbf{(b)} Illustration of Theorem~\ref{theorem4} on the phenomenon of semantic gradients. When minimizing $\mathbf{R}_{\textup{Madry}}(\theta)$, the gradient alignment term, i.e., $\mathbf{R}_{\textup{SCORE}}^{\ell_{1}}(\theta)-\mathbf{R}_{\textup{Standard}}^{\ell_{1}}(\theta)\approx 2\epsilon\cdot\mathbb{E}_{p_{d}(x)}\left[\left\|\nabla_{x}p_{d}(\mathcal{Y}_{d}(x)|x)-\nabla_{x}p_{\theta}(\mathcal{Y}_{d}(x)|x)\right\|_{q}\right]$ keeps decreasing before $\mathbf{R}_{\textup{SCORE}}^{\ell_{1}}(\theta)$ is minimized to be less than $C^{\ell_{1}}$.}
\label{appendixfig:2}
\end{center}
\vspace{-0.cm}
\end{figure}

Recall the discussion in Sec.~\ref{sec41}, actually when $\mathbf{R}_{\textup{SCORE}}^{\ell_{1}}(\theta)=0$, there is $p_{\theta}(y|x)=p_{d}(y|x)$. Thus, in this case we have $\mathbf{R}_{\textup{Madry}}(\theta)=C^{\textrm{KL}}$, where
\begin{equation}
    C^{\textrm{KL}}=\mathbb{E}_{p_{d}(x)}\left[\max_{{x'}\in B(x)}\textrm{KL}\left(p_{d}(y|x)\big\|p_{d}(y|{x'})\right)\right]\textrm{.}
\end{equation}
According to Pinsker's inequality, there is $C^{\textrm{KL}}\geq(C^{\ell_{1}})^{2}/2$. This implies that we should early-stop $\mathbf{R}_{\textup{Madry}}(\theta)$ even earlier at $C^{\textrm{KL}}$. In Fig.~\ref{appendixfig:2} (a), we minimize $\mathbf{R}_{\textup{Madry}}(\theta)$ in training, and find that $\mathbf{R}_{\textup{SCORE}}^{\ell_{1}}(\theta)$ indeed begins to overfit when $\mathbf{R}_{\textup{Madry}}(\theta)\approx C^{\textrm{KL}}$. Nevertheless, Corollary~\ref{corollary1} analytically describes how minimizing $\mathbf{R}_{\textup{Madry}}(\theta)$ affects $\mathbf{R}_{\textup{SCORE}}^{\ell_{1}}(\theta)$.

\subsection{Visualization of Semantic Gradients}
\label{VSG}
In Fig.~\ref{fig:3} and Fig.~\ref{appendixfig:1}, we visualize the semantic input gradients of a ResNet-50 trained by FreeAT on ImageNet. Unlike previous work that directly plots all the perturbation after normalization~\citep{pang2020boosting}, we add up the partial derivatives of three RGB channels for each pixel position, and sort out the top 10\% pixel positions with large values of total derivatives (i.e., affect the objectives the most) in the plots. This way can highlight the shape-based characteristics learned by the robust models. Since $\log p_{\theta}(\mathcal{Y}_{d}(x)|x)$ is a discriminative objective, we also consider $\log p_{\theta}(x,\mathcal{Y}_{d}(x))$ and $\log p_{\theta}(x)$ based on the fact that $\log p_{\theta}(\mathcal{Y}_{d}(x)|x)=\log p_{\theta}(x,\mathcal{Y}_{d}(x))-\log p_{\theta}(x)$, and similarly $\nabla_{x}\log p_{\theta}(\mathcal{Y}_{d}(x)|x)=\nabla_{x}\log p_{\theta}(x,\mathcal{Y}_{d}(x))-\nabla_{x}\log p_{\theta}(x)$. According to \citet{Grathwohl2020Your}, given a classifier $p_{\theta}(y|x)$ with a softmax final layer, i.e.,
\begin{equation}
    p_{\theta}(y|x)=\frac{\exp(f_{\theta}(x)[y])}{\sum_{y'}\exp(f_{\theta}(x)[y'])}\textrm{,}
\end{equation}
where $f_{\theta}(x)[y]$ indicates the $y^{\textrm{th}}$ index of $f_{\theta}(x)$, i.e., the logit corresponding the the class label $y$. Then we can reinterpret the discriminative model as an EBM for the joint distribution $p_{\theta}(x,y)$, formulated as
\begin{equation}
    p_{\theta}(x,y)=\frac{\exp(f_{\theta}(x)[y])}{Z(\theta)}\textrm{; }p_{\theta}(x)=\frac{\sum_{y'}\exp(f_{\theta}(x)[y'])}{Z(\theta)}\textrm{,}
\end{equation}
where $Z(\theta)$ is the unknown normalizing constant. Since there is $\nabla_{x}\log Z(\theta)=0$, we have $\nabla_{x}\log p_{\theta}(x,y)=\nabla_{x}f_{\theta}(x)[y]$ and $\nabla_{x}\log p_{\theta}(x)=\nabla_{x}\log\sum_{y'}\exp(f_{\theta}(x)[y'])$. We can compute $\nabla_{x}\log p_{\theta}(x)$ using the numerically stable operator of Log-Sum-Exp. Note that in the figures, we plot $\nabla_{x}\log p_{\theta}(x,\mathcal{Y}_{d}(x))$ (i.e., the fastest direction of \emph{increasing} $p_{\theta}(x,\mathcal{Y}_{d}(x))$) and $-\nabla_{x}\log p_{\theta}(x)$ (i.e., the fastest direction of \emph{decreasing} $p_{\theta}(x)$).

\subsection{Checking for Gradient Obfuscation}
\begin{figure}[t]
\begin{center}
\vspace{-0.cm}
\includegraphics[width=0.6\columnwidth]{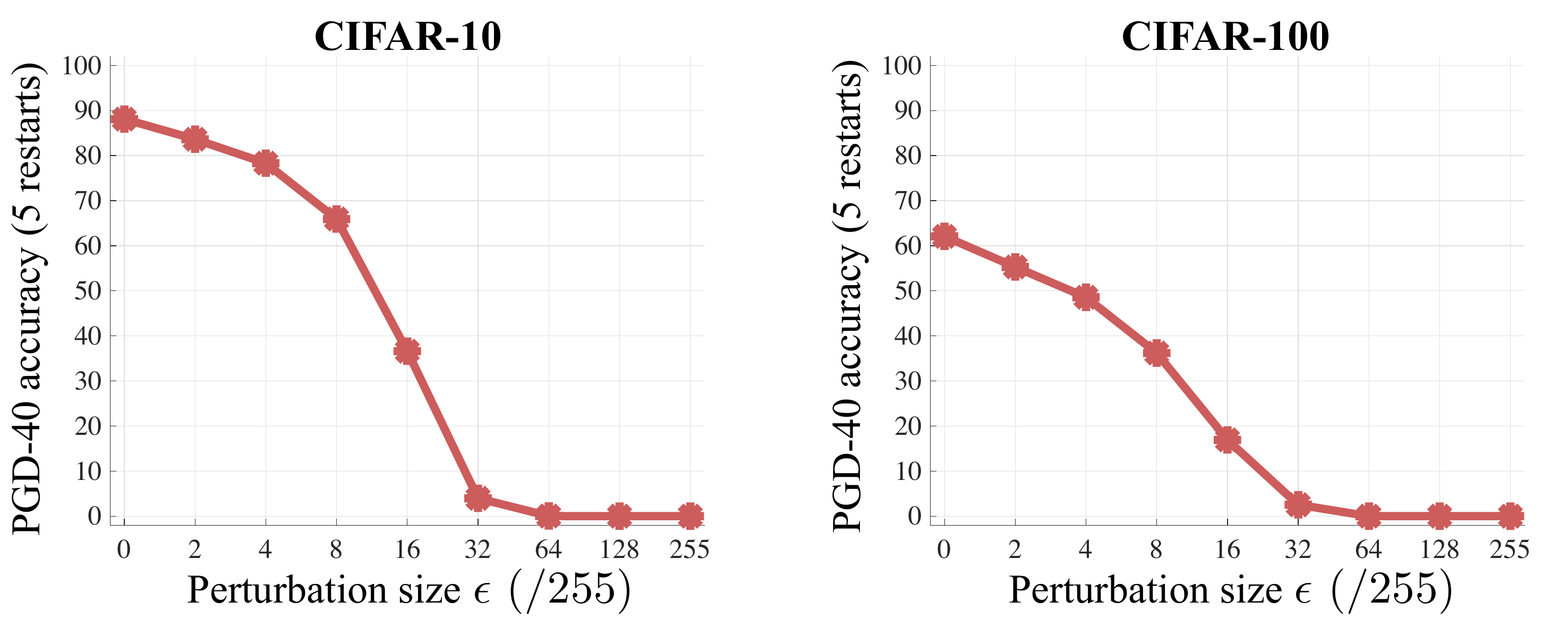}
\vspace{-0.2cm}
\caption{Sanity check on gradient obfuscation. We employ $\ell_{\infty}$ threat model and PGD-40 attacks, with perturbation size $\epsilon$ increasing from $0/255$ to $255/255$. We evaluate the WRN-28-10 models trained by our methods (the SE function) on CIFAR-10 and CIFAR-100.}
\label{appendixfig:3}
\end{center}
\vspace{-0.cm}
\end{figure}

Following the evaluation guide by \citet{carlini2019evaluating}, we perform sanity check for gradient obfuscation~\citep{athalye2018obfuscated}. In Fig.~\ref{appendixfig:3}, we apply PGD-40 attacks with different values of perturbation sizes. We apply five restarts since more restarts only marginally lower the robust accuracy. As seen, on both CIFAR-10 and CIFAR-100, the model accuracy finally converges to zero, indicating that the SE function does not lead to gradient obfuscation.

\section{Clarification on Backgrounds}
This section clarifies some concepts discussed in our papers to avoid ambiguity.

\subsection{Overfitting; Catastrophic Overfitting; Benign Overfitting}
In Sec.~\ref{sec41}, the \emph{overfitting} phenomenon refers to the one observed in multi-step adversarial training~\citep{rice2020overfitting}. In contrast, there is another analogous concept, named \emph{catastrophic overfitting}, which is observed in one-step adversarial training~\citep{wong2020fast}. While we focus on the former overfitting in the sense of a generalization gap, much work focuses on the latter one to alleviate or understand catastrophic overfitting in which the robust accuracy rapidly drops~\citep{andriushchenko2020understanding,li2020towards,vivek2020single,kang2021understanding,zhang2021revisiting,kim2020understanding,golgooni2021zerograd}. Recently, the concept of \emph{benign overfitting} has also drawn attention~\citep{sanyal2021how,chen2021benign}, studying the phenomenon that classifiers memorize noisy training data yet still achieve a good generalization performance.

\subsection{Other Trade-offs in the Adversarial Literature}
This paper focuses on the robustness-accuracy trade-off, which refers to the observations that a model achieves a higher robust accuracy (usually via adversarial training) at the cost of degraded clean accuracy. Nevertheless, several other trade-off phenomena are studied in previous work. For examples, \citet{engstrom2017rotation} show that state-of-the-art $\ell_{\infty}$ robust models turn out to be vulnerable to translations and rotations; \citet{tramer2019adversarial} demonstrate a trade-off in robustness to different types of $\ell_{p}$-bounded (e.g., $\ell_{1}$, $\ell_{2}$, and $\ell_{\infty}$) perturbations; \citet{tramer2020fundamental} present fundamental trade-offs between sensitivity-based and invariance-based adversarial examples.

\subsection{AutoAttack for Evaluation}
In our experiments, we mainly apply AutoAttack for evaluating the robustness of baselines and our methods. We have to clarify that there are many other benchmarks and relevant attacks in the literature~\citep{chen2020rays,dong2019benchmarking,sriramanan2020guided,tang2021robustart}, while recent work also argues that the adversarial examples crafted by AutoAttack are easy to be detected~\citep{lorenz2021robustbench}. Nevertheless, RobustBench\footnote{https://robustbench.github.io/} (based on AutoAttack) is widely recognized and a challenging benchmark, with frequent updates on the state-of-the-art models. Achieving top-rank performance on RobustBench is a compelling evidence for the effectiveness of the proposed method.

\end{document}